\documentclass[10pt,twocolumn,letterpaper]{article}

 \usepackage{iccv}              % To produce the CAMERA-READY version
\definecolor{iccvblue}{rgb}{0.21,0.49,0.74}
\usepackage[pagebackref,breaklinks,colorlinks,allcolors=iccvblue]{hyperref}

\def\paperID{4024} % *** Enter the Paper ID here
\def\confName{ICCV}
\def\confYear{2025}

\usepackage{colortbl}  %彩色表格需要加载的宏包
\usepackage{xcolor}
\usepackage{array} 
\usepackage{epsfig}
\usepackage{graphicx}
\usepackage{amsmath}
\usepackage{amssymb}
\usepackage{multirow}
\usepackage{chemformula}
\usepackage{booktabs}
\usepackage{url}
\usepackage{graphicx}
\usepackage{algorithm}  
\usepackage{algorithmicx} 
\usepackage{algpseudocode}
\usepackage{adjustbox}
\usepackage{caption}
\usepackage{chemformula}
\usepackage{subfloat}
\usepackage{chemformula}
\usepackage{tabularx} 
\usepackage{ragged2e} 
\usepackage{booktabs}
\usepackage{makecell}
\usepackage{booktabs}
\usepackage{wrapfig}
\usepackage[accsupp]{axessibility}
\usepackage{amsfonts}
\usepackage{enumitem}
\usepackage{bbding}
\usepackage{placeins}
\usepackage{xspace}% blackboard math 

\title{4D Visual Pre-training for Robot Learning}

\usepackage{xspace}

\newcommand{\ours}[0]{FVP\xspace}

\author{%
\textbf{Chengkai Hou$^{1}$},  \textbf{Yanjie Ze$^{3}$}, \textbf{Yankai Fu$^1$}, \textbf{Zeyu Gao$^4$}, \textbf{Songbo Hu$^2$},  \textbf{Yue Yu$^2$} \\ \textbf{Shanghang Zhang$^1$,}  \textbf{Huazhe Xu$^{2,3,5}$}\\
$^1$Peking University\quad $^2$Tsinghua University\quad $^3$Shanghai Qizhi Institute \\ \quad $^4$ CASIA\
\quad $^5$ Shanghai AI Lab
}

\begin{document}
%\maketitle

%=============================================================================
% \FloatBarrier
% \begin{figure*}[t]
%     \centering
%     \includegraphics[width=0.97\textwidth]{sec/Figure/teaser.png}
%    \caption{ \textbf{FVP} is a novel 3D point cloud  representation learning pipeline for robotic manipulation. Different from prior works in Contrastive Learning and Masked Signal Modeling,  FVP trains 3D visual representations by leveraging the preceding frame point cloud and employing a diffusion model to predict the point cloud of the current frame. }
%     \label{fig:teaser}
%     \vspace{-1.5em}
% \end{figure*} 
% \FloatBarrier

\twocolumn[
{%
\renewcommand\twocolumn[1][]{#1}
\maketitle
\begin{center}
  \centering
  \begin{minipage}[t]{\linewidth}
    \includegraphics[width=0.99\textwidth]{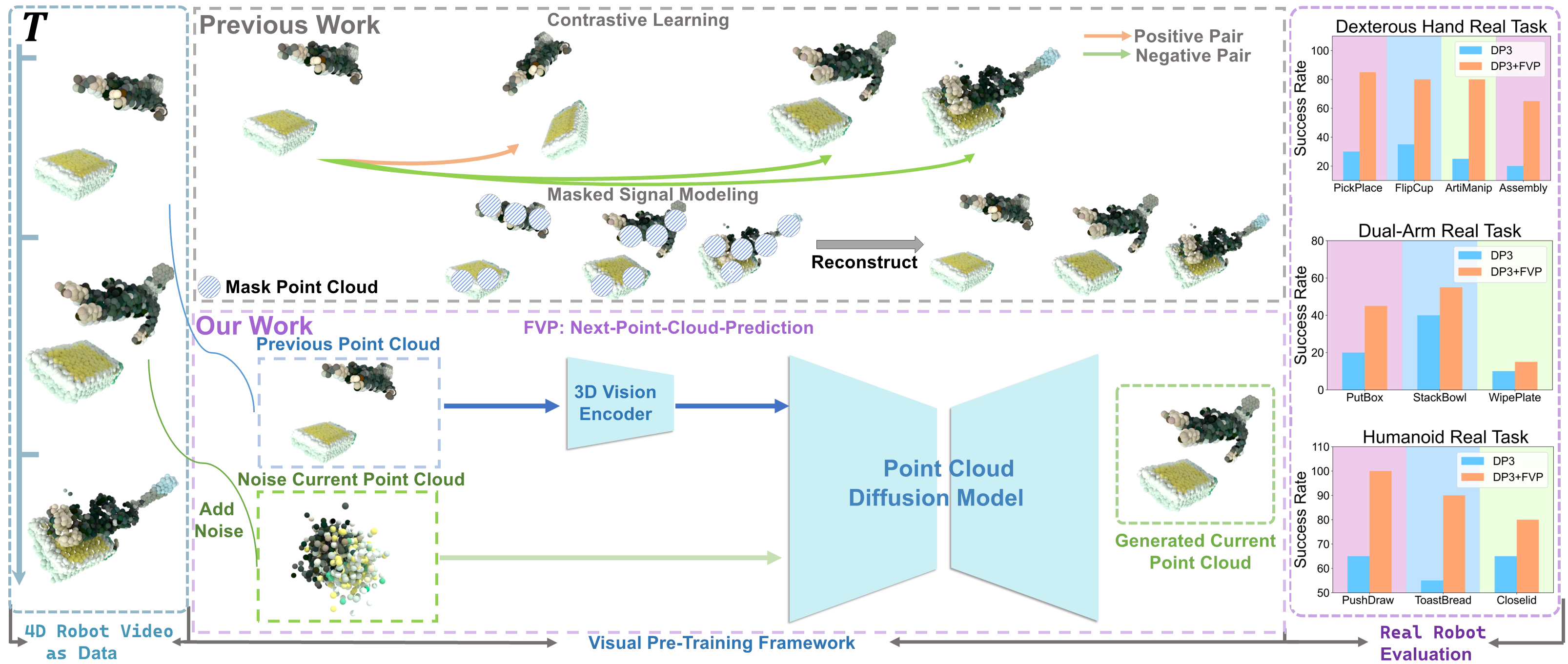}
    % \vspace{-0.6cm}
    {\captionsetup{hypcap=false}  
    \captionof{figure}{\textbf{FVP} is a novel 3D point cloud  representation learning pipeline for robotic manipulation. Different from prior works in Contrastive Learning and Masked Signal Modeling,  FVP trains 3D visual representations by leveraging the preceding frame point cloud and employing a diffusion model to predict the point cloud of the current frame.}
    \label{fig:teaser}}
    % \vspace{+0.1cm}
  \end{minipage}
  % \label{fig:dataset_overview}
\end{center}
}]

\begin{abstract}
    General visual representations learned from web-scale datasets for robotics have achieved great success in recent years, enabling data-efficient robot learning on manipulation tasks; yet these pre-trained representations are mostly on 2D images, neglecting the inherent 3D nature of the world. 
   However, due to the scarcity of large-scale 3D data, it is still hard to extract a universal 3D representation from web datasets. Instead, we are seeking a \textbf{general} visual pre-training framework that could improve all 3D representations as an alternative. Our framework, called \ours, is a novel \textbf{4}D \textbf{V}isual \textbf{P}re-training framework for real-world robot learning. \ours frames the visual pre-training objective as a next-point-cloud-prediction problem, models the prediction model as a diffusion model, and pre-trains the model on the larger public datasets directly. Across twelve real-world manipulation tasks, \ours boosts the average success rate of 3D Diffusion Policy (DP3) for these tasks by \textbf{28\%}. The FVP pre-trained DP3 achieves state-of-the-art performance across imitation learning methods. Moreover, the efficacy of \ours adapts across various point cloud encoders and datasets. 
   Finally, we apply FVP to the RDT-1B, a larger Vision-Language-Action robotic model, enhancing its performance on various robot tasks. Our project page is available at: \url{https://4d-visual-pretraining.github.io/}.
   %, showing its potential in training on large robot datasets.
   %\ze{more exp here}

\end{abstract}

%\keywords{Imitation Learning,  Visual Pre-training, Robotic Manipulation} 

%===============================================================================

\section{Introduction}

Learning generalizable visual representations from large-scale datasets is crucial for robotic tasks~\cite{xiao2022mvp,radosavovic2023robot,radosavovic2023realmvp, nair2022r3m, ze2023vrl3d}.
Currently, robot representation learning is predominantly pre-trained with large-scale 2D images~\citep{xiao2022mvp,radosavovic2023realmvp,nair2022r3m,majumdar2024vc1}. 
However, using 3D point clouds instead of 2D images as visual sources for robotic manipulation has shown efficiency and generalization abilities on real-world robotic tasks~\cite{ze2024dp3, wang2024rise,gervet2023act3d,shridhar2023peract,goyal2023rvt,ze2023gnfactor}.  Thus, we ask: \textit{how can we pre-train for 3D inputs and extract useful representations for robots?}

%\textit{can we provide a \textbf{general} 3D representation for robot manipulation?}

Unlike the abundance of 2D images available on the Internet, 3D point clouds are difficult to obtain from the open web.
Consequently, rather than training a singular visual representation to address multiple robotic tasks, we propose a self-supervised 3D pre-training methodology that is suitable for diverse neural encoders, aimed at enhancing the performance of 3D manipulation tasks.
Due to applying the diffusion model to learn the representations has yielded excellent results in visual tasks~\cite{abstreiter2021diffusionrepresentation,hudson2023soda,wei2023diffusionmae,zheng2023pointpretrain}, we instantiate this idea by employing a straightforward process of iteratively refining the noisy point clouds. Meanwhile, in order to acquire  visual features that  understand the physical environment of robots, we also incorporate the robot action information and the historical frame of robotic point cloud scene into the diffusion process.

Our method, dubbed \ours, frames the learning objective as a \textit{next-point-cloud-prediction} problem and models the prediction network as a conditional diffusion probabilistic model. Notably, \ours directly pre-trains on the robot trajectories (\textit{e.g.}, sequences of observation-action pairs), rendering \ours a general plug-in 4D pre-training module for all 3D imitation learning methods.
\ours first embeds the history frames of the observed point cloud into the latent visual representations using a standard visual encoder such as PointNet++~\citep{qi2017pointnet++}, Point Transformer~\citep{zhao2021point_transformer}, and DP3 Encoder~\citep{ze2024dp3}. Then, conditioning on the 3D visual representations, a modified Point-Voxel Diffusion network~\citep{liu2019pvcnn,zhou2021point_voxel_diffusion} gradually denoises the Gaussian noise into the point clouds of the next frame, as shown in Figure~\ref{fig:teaser}.

% Our pre-training framework has two main advantages.
% First,
% %unlike previous approaches that use diverse images from the Internet to train a strong 2D visual representation, 
% \ours adopts the 3D point cloud collected from real-world robot tasks to pre-train, enabling the learned visual representations to comprehensively understand the real physical world. 
%Second, compared to reconstruction in pixel space like masked image modeling~\cite{xiao2022mvp}, \ours performs future prediction in 3D space, which aggregates the spatial \& temporal point cloud information from different frames along with robot actions and helps robots better understand the downstream tasks.
In contrast to past point cloud pre-training methods such as contrastive learning or point cloud reconstruction, FVP introduces a novel approach by predicting the next frame of point cloud. Traditional methods~\cite{zhang2022pointm2ae,huang2021strl,zhang2023c2p,pang2022pointmae} typically use contrastive learning where point clouds from the same time step are treated as positive pairs and those from different time steps as negative pairs; another approach is to employ point cloud reconstruction by masking portions of the point cloud (see Figure~\ref{fig:teaser}). 
However, FVP leverages the current robot observation predict the subsequent robot observation. Specifically, it enables the visual model to learn to predict the robot's next action based on the current observation. This predictive mechanism allows the visual model to better capture the motion characteristics of the robot, leading to enhanced performance in real-world robotic applications. By focusing on predicting future states, FVP enables more accurate and robust learning of dynamic behaviors—an ability that is critical for robotic tasks.

To demonstrate the effectiveness of \ours, we construct a comprehensive set of tasks comprising 12 simulation tasks and 12 real-world tasks. Simulation tasks are selected from the Adroit~\cite{rajeswaran2017adroit} and MetaWorld~\cite{yu2020metaworld} benchmarks.
In the real-world tasks, the robots used include single-arm robots equipped with grippers and dexterous hands, dual-arm robots, and humanoid robots.
For the \textbf{Simulation} tasks, regardless of whether in-domain or out-of-domain datasets are used for pre-training, FVP-pretrained DP3 achieves the state-of-the-art performance on various simulator tasks.  Specifically, it improves average task accuracy by \textbf{17\%} when using in-domain datasets and by \textbf{24.7\%} when using out-of-domain datasets.
For the \textbf{Real} tasks, we observe that \ours could achieve \textbf{15\%$\sim$55\%} absolute improvements when built upon the state-of-the-art 3D imitation learning methods, \textit{e.g.,} DP3~\citep{ze2024dp3} and RISE~\citep{wang2024rise}, and largely surpass other 2D methods such as ACT~\citep{zhao2023act} and Diffusion Policy~\citep{chi2023diffusionpolicy} (see Figure~\ref{fig:teaser}). 
Moreover, we show that \ours could improve over different 3D encoders including DP3 Encoder~\citep{ze2024dp3}, PointNet++\citep{qi2017pointnet++}, and Point Transformer~\citep{zhao2021point_transformer}, showing the potential in pre-training on large-scale datasets. 
{Then, the visual models pre-trained by \ours are leveraged in  the Vision-Language-Action Robotic models (VLA model), specifically RDT-1B~\cite{liu2024rdt}.
We demonstrate through real-world tasks involving both single-arm and dual-arm robots that 3D point cloud input can effectively improve the efficiency and generalization of RDT models. 
Additionally, utilizing the FVP pre-trained 3D encoder on the RoboMind dataset enhances the RDT-1B model's abilities in several key areas: spatial perception, language understanding, and task generalization. We are committed to releasing the code.

\section{Related Work}

\noindent\textbf{Visual representations for robotics.}  
In recent years, the field of visual representations for robotics has seen significant advancements, driven by the need for robots to better understand and interact with their environments. 
% One line of work is image-based visual representation
%\ze{what is "image based visual representation", seems weird. it is maybe just 2D visual representation}.
Most works use 2D visual representations for robot control, learning from large-scale web datasets such as ImageNet~\citep{deng2009imagenet,shah2021rrl} and Ego4D~\citep{nair2022r3m,grauman2022ego4d,xiao2022mvp,radosavovic2023realmvp}. Among them, R3M~\citep{nair2022r3m} explores Time Contrastive Learning and Video-Language Alignment to train a universal representation for robots. MVP~\citep{xiao2022mvp} follows the masked autoencoder paradigm and learns from Ego4D videos. VC-1~\citep{majumdar2024vc1} scales up the model size and dataset in MVP. 
% In-domain data uses the data from target environment and task for training the representation. 
Recently, learning visuomotor policies from point clouds instead of images has shown great potential~\citep{shridhar2023peract,ze2023gnfactor,ze2024dp3,wang2024rise}, while a universal pre-training paradigm on point clouds for robotics is yet to be studied.

\noindent\textbf{Visual imitation learning} provides an efficient way to teach robots human skills from human demonstrations and the learned skills could be more easily deployed in the real world compared to state-based methods~\citep{ze2023vrl3d,zhao2023act,chi2023diffusionpolicy,ze2024dp3,shridhar2023peract}.
% In the imitation learning community, most previous solutions are designed for state-based settings, where the states/observations are low-level perceptive vector inputs~\cite{liu2020energy,liu2022plan,garg2021iqlearn,kostrikov2018discriminator}.
% However, these low-level information vectors  are insufficient to assist robots in accomplishing complex real tasks.
% Currently,  many researchers propose many methods to achieve end-to-end robot manipulation by capturing 2D image information, such as.
Nonetheless, 2D imitation learning methods such as ACT~\cite{zhao2023act} and Diffusion Policy~\cite{chi2023diffusionpolicy} are sensitive to camera positions and often fail to capture 3D spatial information about the objects in the environments, which highlights the necessity of 3D observations.
% Thus, it is necessary to explore 3D visual information to implement robot control. 
ACT3D~\cite{gervet2023act3d} explores the features of multi-view RGB images with a pre-trained 2D backbone and lifts them in 3D to predict the robot actions.
DP3~\cite{ze2024dp3} utilizes lightweight encoders to extract point cloud features, which are then fed into a diffusion model to predict the robot trajectory. Rise~\cite{wang2024rise} adopts a more complex structure, including sparse convolutional networks and transformers, to encode the point cloud into point tokens and then uses these tokens to predict actions.

\noindent\textbf{Diffusion models for robotics.} 
Diffusion models are a kind of generative models that learn a denoising process by the diffusion process. 
% that transform random noise into a probability distribution of data samples.
They have been gaining significant popularity in the past few years due to their excellent performance in image generation~\citep{ho2020ddpm,song2020score,song2020ddim,rombach2022stable_diffusion} and point cloud generation~\citep{zhou2021point_voxel_diffusion, yang2024visual,mersch2022self}. Due to the expressiveness of diffusion models, they have been applied in robotics recently, such as reinforcement learning ~\citep{sridhar2024memoryconsistent,ajay2023conditional}, imitation learning ~\citep{florence2021implicit,nuti2023extracting,yan2024nerfuser,qin2022dexmv,urain2023se3diffusionfields,chi2023diffusionpolicy,ze2024dp3}, reward learning ~\citep{ho2020ddpm,mandlekar2021matters,huang2023diffusion_reward}, grasping ~\citep{6386109,simeonov2023shelving,shafiullah2023bringing}, and motion planning~\citep{rajeswaran2018learning}. Different from these works, this work provides a visual pre-training framework for robotics that is based on diffusion models.

\section{Method}

In this section, we describe the details of our proposed \textbf{4}D \textbf{V}isual \textbf{P}re-training (\textbf{\ours}). We begin by giving an introduction to diffusion models and then describe how \ours pre-trains 3D visual representations and applies the pre-trained representations for downstream robotic manipulation tasks.

%\begin{figure*}[t]
%    \centering
%    \includegraphics[width=1.0\textwidth]{sec/Figure/method.pdf}
%    \caption{ \textbf{Overview of 4D Diffusion Pre-training (\ours).}
%    \ours mainly consists of two parts: a 3D visual encoder and a point cloud diffusion model. The 3D visual encoder transforms the point cloud at time step $t$ into the latent visual representation, and the diffusion model uses the latent visual representation and robotic actions to predict the point cloud at step time $t+1$. During the policy learning stage, we train the pre-trained visual model and the policy backbone jointly.}
%    \label{fig:basemethod}
%    \vspace{-1.5em}
%\end{figure*} 

\subsection{Diffusion Models Revisited}
We first give a brief introduction to the denoising diffusion probabilistic model which generates 3D point clouds through denoising process from random Gaussian noises~\citep{zhou2021point_voxel_diffusion,ho2020ddpm,song2020score,song2020ddim}.  %The denoising process converts a noise distribution  into the meaningful point cloud 
%distribution.
During training, diffusion models add a series of noises to the original point cloud $X_0$ as input, represented as $X_{T}$. The process of adding noise, \textit{e.g.}, the diffusion process, is modeled as a Markov chain~\cite{jarzynski1997equilibrium}:
\begin{equation} 
\begin{split}
    q(X_{1:T}|X_0)&=\prod_{t=1}^T q(X_{t}|X_{t-1}), \\
    q(X_{t}|X_{t-1}) &=\mathcal{N}(X_t;\sqrt{1-\beta_t}X_{t-1},\beta_t \mathbf{I}).
\end{split}
\label{eq:ddpm1}
\end{equation}
where $T$ denotes the number of steps and $q({x_t}|{x_{t-1}})$ is a Gaussian transition kernel, which gradually adds noise to the input with a variance schedule $\{ \beta_t \}_{t=0}^{T}$. 
Thus, by progressively inferring the point cloud distribution, we can obtain:
\begin{equation} 
    q(X_t|X_0) = \sqrt{\bar{\alpha}_t} X_0 + \epsilon \sqrt{1-\bar{\alpha}_t},
\end{equation}
where $\alpha_t = 1 - \beta_t$, $\bar{\alpha}_t = \prod_{s=0}^{t} \alpha_s$, and $\epsilon \sim \mathcal{N}(0, \mathbf{I})$. In order to generate a recognizable object, we learn a parametrized reverse process, which denoises the noise distribution $q(X_{T})$ into the target distribution $q(X_{0})$.
To achieve the reverse process, we utilize the network $\epsilon_{\theta}$ to learn the reverse process $q(X_{t-1}|X_{t})$. 
$\epsilon_{\theta}$: $\mathbb{R}^{ N\times 3}\to \mathbb{R}^{ N\times 3}$ is a diffusion model which assigns the points from Gaussian noise ball  
into the optimal location. Specially, at each step, we use the network to predict the offset of each point from current location and through each step iterates,  the noisy point will arrive in the ideal position. 
Thus, the network is required to output the added noise $\epsilon$ at the most recent time step $T$ to denoise. We use the $L_{2}$ loss $\mathcal{L}$ between the predicted noise and the ground truth $\epsilon \in \mathbb{R}^{ N\times 3}$ to optimize the network:
\begin{equation}
\mathcal{L} = E_{\epsilon \sim \mathcal{N}(0, \mathbf{I})}\left[\|\epsilon - \epsilon_{\theta}(X_t, t)\|_2^2\right]
\end{equation}

At the inference time,  we reverse the diffusion process that denoises the point cloud with a standard 3D Gaussian distribution $X_T \sim N(\textbf{0}, I_{3N})$   into a recognizable sample $X_{0}$ iteratively.
\subsection{4D Visual Pre-training on 3D Visual Representations}

\noindent\textbf{Demonstration collection.}
%We adopt the robotic datasets to pre-train the visual model in \ours.
%We collect the expert demonstrations using the RealSense camera for each task. %These demonstrations include the 3D point cloud video, robotic action and state information. 
%Each task demonstrations are used to pre-train its own visual encoder.
%To achieve the vision model 
To pre-train 3D visual representations for downstream robotic manipulation tasks,  we access the demonstrations $\mathbf{X} = \{x^{\mathit{0}},x^{\mathit{1}},\dots,x^{\mathit{T}} \}$ collected from the real-world robotic tasks, where each trajectory contains $\mathit{T}$ frames of observation-action pairs $x^{t}=(o^t, a^t)$. The observation $o^t$ is the 3D point cloud at time $t$ and the action is $a^t$ the robot joint position at time $t$.
Each task demonstrations are used to pre-train its own visual encoder.
FVP is also applicable for out-of-domain pre-training using publicly available robot datasets such as Robomind, as long as they contain complete point cloud information for robotic manipulation.
% These tasks can be demonstrated using either datasets we have collected in-domain or publicly available robot data sets, such as the robomind dataset~\cite{wu2024Robomind}.
 %The learning objective of \ours is to predict the future point cloud based on the point cloud and the corresponding action of the current frame. 
 %\ours first extracts the latent feature from the previous point cloud  $o^{\mathit{t-1}}$ and robotactions to help the diffusion model to convert  the noise point cloud distribution $x^{\mathit{t}}_{T} $ into the target one.\ze{the logic seems to be very confusing}
%Initially, we integrate the preceding point clouds $x^{\mathit{t-1}}$ and robotic action states into latent features $\mathbf{z}$ using visual extraction. 
%Subsequently, prior to the denoising phase, we project these features $\mathbf{z}$ onto the noisy point cloud $x^{\mathit{t}}_{T}$, associating a neural feature with each point. 
%Finally, each input point, along with its corresponding the neural feature, is diffused as the novel samples (see Figure ). In the subsequent section, we will elaborate on the generation process of the latent representation $\mathbf{z}$ and its attachment to the point cloud $x^{\mathit{t}}_{T}$.

%Meanwhile, we integrate robot action information into the pre-training of the vision model, enabling a deeper understanding of the physical scenarios of robot operations.

%\paragraph{Encoder.}

\noindent\textbf{Extracting 3D visual representations.} \ours  encodes the previous frame's point cloud $o^{t-1}$ into a latent representation $\mathbf{z}$, which is to guide the diffusion model to predict the future frame point cloud $o^{t}$  (Figure~\ref{fig:teaser}). 
The visual encoder could be implemented as any type of general 3D encoders, such as PointNet++~\citep{qi2017pointnet++}, Point Transformer~\citep{zhao2021point_transformer}, DP3 Encoder~\citep{ze2024dp3}, and RISE Encoder~\citep{wang2024rise}.
% The latent representation $\mathbf{z}$ consists of the visual feature, generated by encoding the previous frame of point cloud $o^{t-1}$ through a visual model and the  low-level dimensional action vector generated by compressing the robot's action information $a^{t-1}, a^{t}$ through a lightweight MLP network.
% For the visual feature,  the 3D visual encoder $f_{\theta}(\cdot)$ encompasses not only the standard point cloud networks like PointNet++~\cite{qi2017pointnet++}, but also 3D perception networks utilized in imitation learning~\cite{wang2024rise,ze2024dp3}.
% We utilize the 3D visual encoder such as DP3 Encoder~\citep{ze2024dp3} and Point Transformer~\citep{zhao2021point_transformer} to encode the preceding frame of point cloud $o^{t-1} $ into the visual representation $\mathbf{z}_v = f_{\theta}(o^{t-1}), \mathbf{z}_v \in \mathbb{R}^{ N\times C_{v}}$.
%For the low-level dimensional action vector, 
% And we employ a lightweight multi-layer perceptron $f_{\beta}(\cdot)$ to embed the previous and current robotic action information $a^{t-1}, a^{t}$ into a low-dimensional action vector $\mathbf{z}_{a}$. The low-dimensional action vector $\mathbf{z}_a \in \mathbb{R}^{ N\times C_{a}}$ is denoted as $z_{a} = f_{\beta}(a^{t-1}\oplus a^{t})$, where $\oplus$ denotes the concatenation  of feature dimensions.
The latent representation $ \mathbf{z} \in \mathbb{R}^{ N\times C_{v}}$, where $N$ is the number of point clouds, $C_{v}$ are the feature dimensions of point clouds. 

%\ze{this paragraph is not very straightforward.}
%In the \ours approach, we involve both the gradually sparse features $\mathbf{z}_{s}$ of point clouds processed through three layers of MLP and the dense features $\mathbf{z}_{d}$ processed through the entire vision model from DP3 in the pre-training process.
%The sparse feature is denoted as $\mathbf{z}{s} = f_{\theta}(x^{t})$, where $\mathbf{z}_{s} \in \mathbb{R}^{ N\times C{s}}$, and the dense feature is represented as $\mathbf{z}_{d} = f_{\beta}(f_{\theta}(x^{t}))$, where $\mathbf{z}_{d} \in \mathbb{R}^{C{d}}$.
%In pursuit of dimensional symmetry between sparse and dense features, we iteratively duplicate the dense feature vector $\mathbf{z}_{d}$ N times, thus yielding a resultant feature vector $\mathbf{z}_{d} \in \mathbb{R}^{N \times C_{d}}$. 
%Meanwhile, we integrate robot action information into the pre-training of the vision model, enabling a deeper understanding of the physical scenarios of robot manipulation. We embed the  robot action information $r$ from both time steps $t$ and $t-1$ into the visual features through an MLP network. Thus, the latent feature $\mathbf{z}$ combines with the repeated dense features $z_{d}$ with the sparse features $z_{s}$ and robot action features $r_{f}$, which is denoted as $\mathbf{z} = z_{d}\oplus z_{s}\oplus r_{f}, \mathbf{z} \in \mathbb{R}^{ N\times C}$, where $C$ is the number of  feature channels and $\oplus$ means the combination of feature dimensions. 

\noindent\textbf{Generating future point cloud.} Conditioning on the latent representation $\mathbf{z}$, our point cloud diffusion model denoises the random Gaussian noise into the future point cloud.
%\ze{to add more details and make it simpler to read.}
{In particular, we project the latent representation $\mathbf{z}$ onto the current frame of point cloud with added noise $o^{t}_{T}$, $T$ represents the number of added noisy steps. The input point cloud of the diffusion model is changed from $o^{t}_{T}\in \mathbb{R}^{ N\times3}$ to $o^{t}_{T,+} \in \mathbb{R}^{ N\times (C_{v}+3)}$. 
$\epsilon_{\theta}$ is now a new function: $\mathbb{R}^{ N\times (C_{v}+3)}\to \mathbb{R}^{ N\times 3}$ which predicts the noise $\epsilon$ from the attached point cloud $o^{t}_{T,+} = [o^{t}_{T}, \mathbf{z}]$.}
Thus,  the optimization of the loss function $\mathcal{L}$ for the neural network   $\epsilon_{\theta}$ is transformed as: 
\begin{equation}
\mathcal{L} = E_{\epsilon \sim \mathcal{N}(0, \mathbf{I})}\left[\|\epsilon - \epsilon_{\theta}(o^{t}_{+,T}, T)\|_2^2\right]
\end{equation}

%\ze{where is the formula of the training objective? }
%Formally, let  $\mathbf{z} \in \mathbb{R}^{ N\times C}$ be the feature consists of the visual feature obtained from the previous point cloud $x^{t-1}$ encoded by the visual model and the robot action embedding information, where $C$ is the number of  feature channels. 
%The diffusion model, denoted by $\epsilon$, refines the current point cloud $x^{t}$ by adding spherical Gaussian ball noise, which is fused with the latent feature $\mathbf{z}$ to the recognizable object. This transformation can be represented as $\mathbb{R}^{ N\times (C+3)}\to \mathbb{R}^{ N\times 3}$.

%\subsection{\ze{Downstream Tasks}}

%\ze{describe how you apply the pre-trained visual representations in downstream tasks}

%3D imitation learning, such as RISE~\cite{wang2024rise} and
%DP3~\cite{ze2024dp3}, can rapidly and accurately inference robotic  trajectory  in the real-world robotic task. To 
\noindent\textbf{Downstream robotic tasks.} After obtaining the pre-trained 3D visual representations, we apply them in downstream real-world robotic manipulation tasks.
Given the collected expert demonstrations,  we train 3D visuomotor policies such as RISE~\cite{wang2024rise} and DP3~\cite{ze2024dp3}, which adopts point clouds as input from time step $t$ and predict robot joint positions for time step $t+1$. We directly replace the original visual representations with the pre-trained ones and fine-tune the visual representations and the policy backbone in an end-to-end manner during training.

\subsection{FVP Visualization}

\begin{figure}[h]
 \centering
 \includegraphics[width=0.45\textwidth]{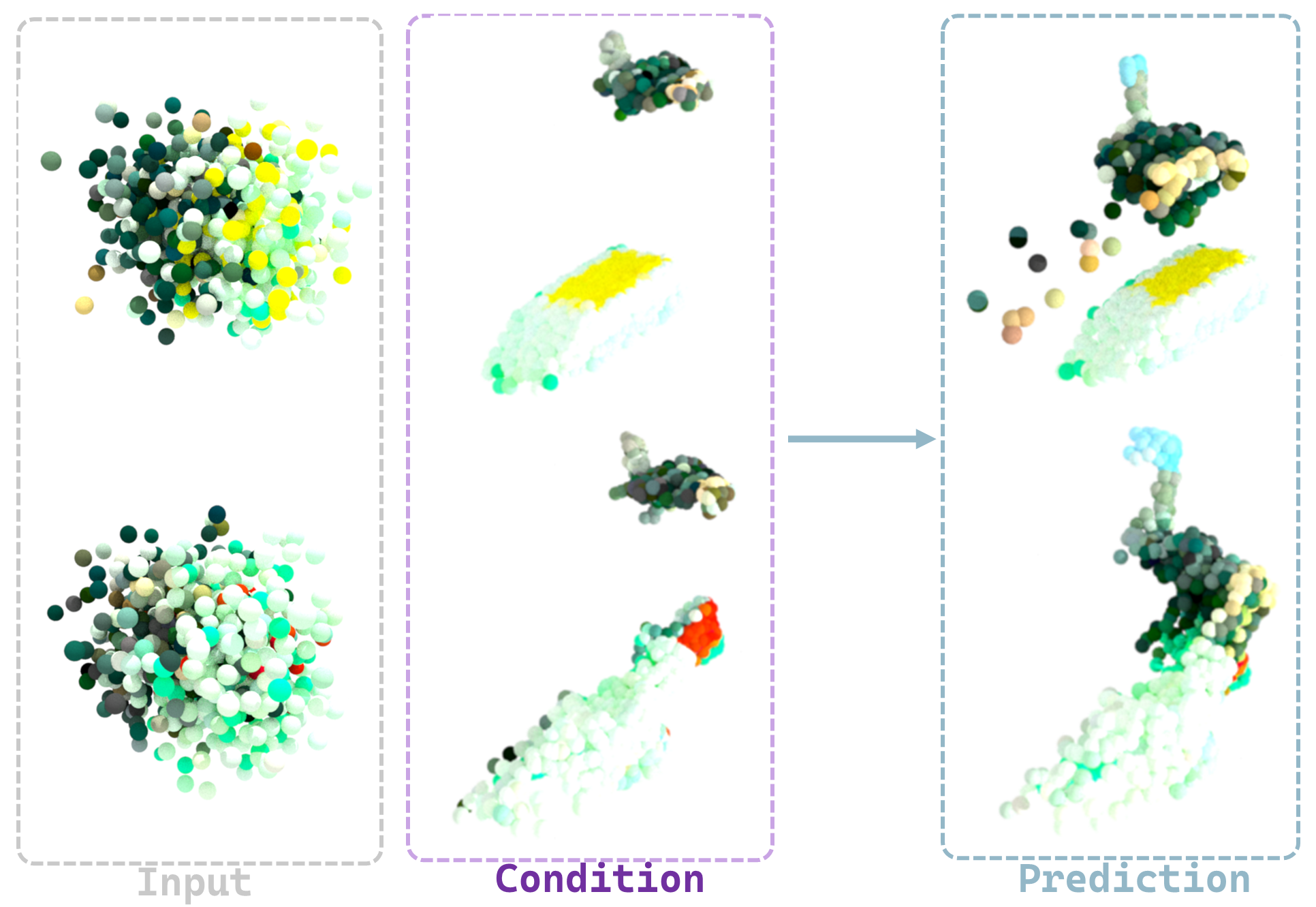}
 \caption{\textbf{Visualization of the  next-point-cloud prediction.} We present a comparison between predictions and conditions along the timeline of the ``Assembly'' task.}
 \label{fig:vis}
 \vspace{-1em}
 \end{figure}

Figure~\ref{fig:vis} illustrates the visualization results of the FVP model predicting the next-frame point cloud. It can be observed that the predicted point cloud distribution closely resembles the ground truth distribution of the subsequent frame. This similarity indicates that FVP has effectively learned temporal dynamics and spatial features that are essential for predicting future states. By capturing the underlying structure and motion patterns of the point cloud sequence, the learned representations demonstrate strong predictive capabilities.

This property is particularly valuable for imitation learning tasks, where actions are generated based on current observations. The ability of FVP to produce feature representations that enable accurate next-frame prediction suggests that these features are well-suited for deriving sequential robotic actions in a manner consistent with the natural evolution of the environment. In essence, FVP provides a robust and predictive feature space that supports effective policy learning from raw sensory inputs.

% We observe in our experiments (see Table~\ref{tab:ablation}) that fintuning helps reduces the gap information between the visual encoder and policy prediction network. 
%Table~\ref{tab:ablation} shows that during the training process of DP3, fine-tuning the visual model pre-trained by \ours yields a \textbf{\%} higher average success rate across the four tasks compared to freezing the pre-visual.
%\ze{more specific by exp}.

\section{Simulation Experiment}
In our experiment, we aim to investigate how the pre-trained visual representations adopted by \ours can be utilized for downstream robotic simulation and real-world manipulation tasks.
As the discrepancy between simulation environments and real-world scenarios diminishes, some standardized simulation benchmarks can serve as effective tools to validate the efficacy of \ours.  
Therefore, in this section,  we evaluate the performance of FVP on simulation tasks from the ``Adroit" and ``Metaworld" benchmarks. 

\subsection{Simulation Benchmark}

\textbf{Adroit.} Adroit~\cite{rajeswaran2017adroit} introduces a set of dexterous manipulation tasks that serve as a benchmark for assessing the capabilities of deep reinforcement learning in controlling a 24-degree-of-freedom hand. The tasks include object relocation, where a ball must be moved to a randomized target location; in-hand manipulation, requiring the repositioning of a pen to match a target orientation; door opening, involving the undoing of a latch and swinging the door open; and tool use, specifically hammering a nail into a board with variable nail positions.

\noindent\textbf{MetaWorld.} MetaWorld~\cite{yu2020metaworld} is a comprehensive benchmark that encompasses 50 diverse simulated robotic manipulation tasks. These tasks are designed to challenge and evaluate the capabilities of meta-reinforcement learning and multi-task learning algorithms in acquiring new skills efficiently. The tasks involve a range of actions such as reaching, pushing, grasping, and placing objects, as well as more complex maneuvers like opening doors, windows, and drawers, turning dials, and inserting pegs into holes. 

%\noindent\textbf{DexArt.}  DexArt~\cite{bao2023dexart} is a benchmark for evaluating generalizable dexterous manipulation skills with articulated objects using point cloud observations. It defines complex tasks such as turning on a faucet, lifting a bucket, opening a laptop lid, and operating a toilet lid, which require a multi-fingered robot hand to manipulate diverse articulated objects. 

\subsection{Evaluation Detail}
The primary objective of FVP is to provide a novel pre-training method to enhance the performance of 3D imitation learning. To this end, our main baselines are several 3D/4D visual pre-training  methods. Additionally, we also compare FVP with 2D pre-training visual models  in terms of their enhancement of imitation learning. 
Meanwhile, to validate the effectiveness of FVP, we employ both in-domain and out-of-domain datasets for pre-training. 
The out-of-domain datasets contain all tasks within the current benchmark, which also include the tested tasks.
For example, for the ``Adroit'', the in-domain dataset consists of datasets for each individual task (``Hammer'', ``Door'', ``Pen''), while the out-of-domain dataset comprises the sum of all tasks datasets on the ``Adorit''.

%\subsection{Evaluation Metric}
Following the DP3 testing pipline, we run 3 seeds for each experiment
with seed number 0, 1, 2. For each seed, we evaluate 20
episodes every 200 training epochs and then compute the
average of the highest 5 success rates. We report the mean
and std of success rates across 3 seeds.

\subsection{Experiment Results}

In Figure~\ref{table:simulation}, we demonstrate the performance of different baselines pre-trained on in-domain and out-of-domain datasets on DP3~\cite{ze2024dp3}.  
We can observe that when pre-training on the in-domain dataset, FVP exhibits an average improvement in the success rate of 16.9\% on the Adorit  and  the Metaworld benchmarks. When FVP adopts the out-of-domain datasets to pre-train the vision encoder, DP3 pre-trained by FVP demonstrates a significant improvement in task success rates on the Adorit  and  Metaworld benchmarks, especially in some difficult tasks (such as Hand Insert and Pick Out of Hole Hand Insert Disassemble).
Thus, we can conclude that FVP demonstrates a more effective ability to improve the success rates of tasks in simulation compared to other pre-training methods, regardless of whether pre-training is conducted on small batches of in-domain datasets or large number of out-of-domain datasets. 
Meanwhile, we evaluate the performance of DP3~\cite{ze2024dp3}, pre-trained with FVP, against 2D imitation learning utilizing a pre-trained vision backbone in Figure~\ref{table:simulation}. Despite being pre-trained on datasets exceeding size 300M, the performance of MVP and R3M in enhancing the success rate of tasks when applied to \textbf{D}iffusion \textbf{P}olicy is inferior to that of FVP pre-trained on in-domain/out-of-domain data in 3D imitation learning. 

\begin{figure*}[h]
\centering
\includegraphics[width=1\textwidth]{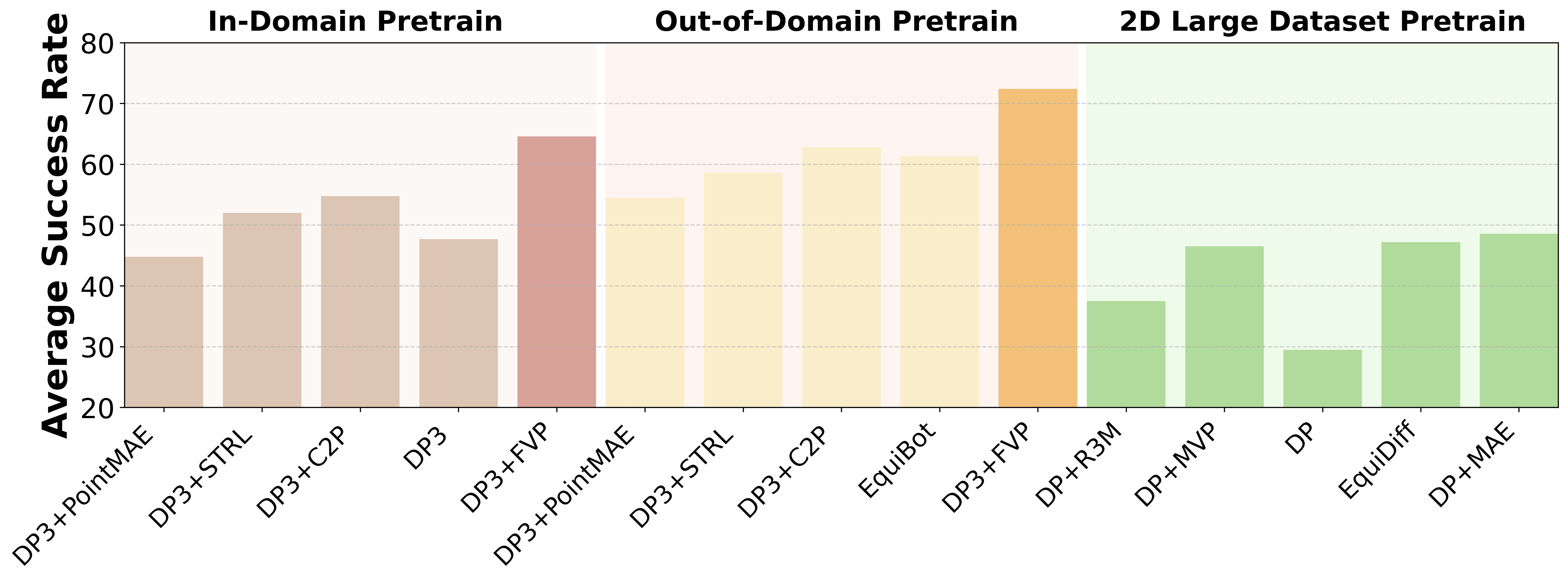}
\vspace{-2em}
\caption{\textbf{Comparing \ours with more baselines in simulation.} We include various 3D pre-training methods, various 2D pre-training methods, and variants of Diffusion Policy such as EquiBot~\citep{yang2024equibot} and EquiDiff~\citep{wang2024equivariant}. }
\label{table:simulation}
\vspace{-1em}
\end{figure*}

\section{Real-world Experiment}

Currently, 3D imitation learning gains widespread application in enabling various types of robots to execute real-world tasks. In this section, we systematically evaluate the extent to which \ours enhances the performance of single task imitation learning and vision-language-action large model(VLA model) in practical tasks. Specifically, we assess the effectiveness of \ours in improving task success rates and robustness across different robotic platforms, including the \textbf{UR5} single-arm robot with  a robotic arm gripper and 16-Dof Leap Hand with four fingers, the \textbf{AgileX} dual-arm robot and the \textbf{TianGong} humanoid robot.

\begin{figure*}[h]
\centering
\includegraphics[width=1\textwidth]{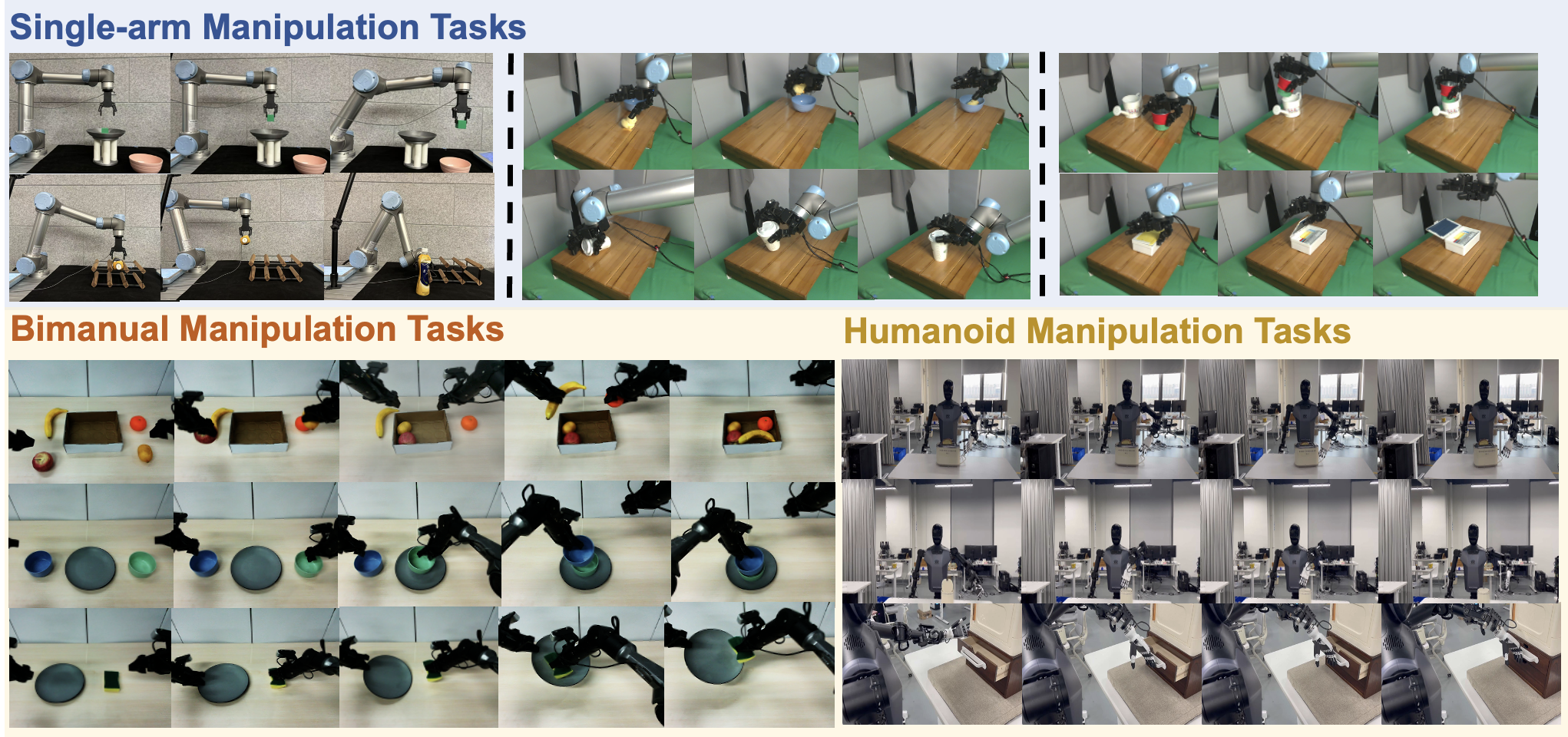}
\vspace{-2em}
\caption{\textbf{Visualization of our real-world tasks.}
For each task, we show several steps to understand the task process.}
\label{fig:task}
\vspace{-1em}
\end{figure*}

%In our experiments, we aim to study how the pre-trained visual representation adopted by \ours can be used for downstream real-world robotic manipulation tasks.
%Firstly, we explore if \ours can boost the performance of imitation learning methods such as DP3~\citep{ze2024dp3} and RISE~\citep{wang2024rise}.
%Secondly, we show that 3D visual representations learned by \ours exhibit better performance compared to other 2D and 3D representations such as R3M~\citep{nair2022r3m} and PointMAE~\citep{zhang2022pointm2ae}.
%Thirdly, we validate the effectiveness of \ours across different visual encoders to improve their performance in real-world robotic tasks.
%Additionally,  we indicate that \ours utilizes the out-of-domain datasets for pre-training the vision model, which can still improve its effectiveness on the real-world robotic tasks.
%Finally, we conduct ablation studies to reveal the key components in \ours and explore the correlation between the number of training expert demonstrations and the effectiveness of \ours.

\subsection{Experiment Setup}
\textbf{UR5 single-arm robot setup.}
We use the UR5 robotic arm equipped with a gripper for real-world robotic tasks. Our visual observations including images and point clouds are collected by one Intel RealSense \textit{L515} RGB-D camera. The camera is placed in the northeast corner of the console, which is approximately 120cm by 60cm in size. For a thorough evaluation of our approach, we design two real-world tasks:
 \begin{itemize}[leftmargin=*]
     \item \textbf{PickSquare}, where the robot picks up the green square and places it in the bowl.
     \item  \textbf{PlaceBottle},  where the robot grabs the bottle and places it on the table.
 \end{itemize}
Then, we equip a UR5 single-arm with a LeapHand dexterous hand as the end effector instead of a gripper, and then we design four tasks to evaluate the effectiveness of FVP. These tasks are explained as follows:
 \begin{itemize}[leftmargin=*]
     \item \textbf{PickPlace:} The dexterous hand picks up a toy chicken and places it into a blue bowl.
     \item  \textbf{FlipCup:} The dexterous hand reaches a cup lying on the table and upright it.
     \item  \textbf{Assembly:} The dexterous hand reaches and grasps a cylindrical cup, lifts it up and inserts it into a kettle.
     \item  \textbf{ArtiManip:} The dexterous hand lifts the lid of a box using its thumb and gently opens it.
 \end{itemize}
 
%\begin{figure}[h]
%\centering
%\includegraphics[scale = 0.5]{sec/Figure/ur5.jpg}
%\caption{\textbf{ UR5 robot setup.} We use a UR5 robot with a parallel gripper to conduct the experiments.}
%\label{fig:ex}
%\end{figure}
\noindent\textbf{AgileX dual-arm robot setup.}
Since many operational tasks in human reality require dual-arm coordination to complete, and dual-arm coordination can achieve higher task efficiency. In our paper, we use the AgileX Cobot Magic~\citep{cobot_magic} dual-arm robot setup designed based on Mobile ALOHA~\citep{fu2024mobile} to perform actual dual-arm tasks to validate the effectiveness of FVP. Additionally, we use the  Intel RealSense 
 \textit{L515} RGB-D camera to record  visual information during task execution. We  provide a detailed description of each dual-arm manipulation task:
 \begin{itemize}[leftmargin=*]
     \item \textbf{PutBox:} Both the left and right arms move the fruits from the table into the box.
     \item  \textbf{StackBowl:} The dual arms stack two bowls on top of each other, with each arm controlling one bowl.
     \item  \textbf{WipePlate:} The left arm holds the sponge and clean the plate picked by the right arm.
 \end{itemize}
\noindent\textbf{TianGong humanoid robot setup.}
We use the built-in cameras of TianGong humanoid robot~\cite{x-humanoid} to collect visual information from real-world task scenarios, including 3D point clouds and 2D images.
Simultaneously, we collect proprioceptive data, such as joint positions and actions, from the upper body of the TianGong humanoid robot.
The upper body of the TianGong robot has 30 degrees of freedom (DoF), distributed across its head, arms, waist, and hands.
Specifically, the head has three degrees of freedom, each arm contains seven degrees of freedom, each dexterous hand has six degrees of freedom, and the waist has one degree of freedom.
To evaluate the performance of \ours in humanoid robots, we design three real-world tasks:
\begin{itemize}[leftmargin=*]
     \item \textbf{PushDraw:} The humanoid robotic arm  pushes in a drawer.
     \item  \textbf{ToastBread:} The humanoid robotic arm starts the toaster to bake bread.
     \item \textbf{Closelid:} The humanoid robot arm closes the garbage lid.
 \end{itemize}
%To evaluate the performance of FVP on humanoid robots, we design four real-world tasks, which are visualized in Figure~\ref{fig:task},
%\begin{itemize}[leftmargin=*]
%     \item \textbf{PullDraw:} The humanoid robotic arm  pull out a drawer.
%     \item  \textbf{:} The humanoid robotic arm close the garbage lid.
%     \item \textbf{Toasting Bread:}, where the robot picks up the computer docking station, places it in the drawer, and then pushes the drawer.
 %\end{itemize}

%\begin{figure}[h]
%\centering
%\includegraphics[scale = 0.3]{sec/Figure/renxing.png}
%\caption{\textbf{KUAVO Humanoid  data collection.} We use Meta Quest3 to project the trajectories of human hand and waist movements onto the humanoid robot's operations, enabling the robot to perform real-world tasks. Meanwhile, the RealSense L515 camera and the Humanoid robot continuously record visual and proprioceptive information.}
%\label{fig:ex}
%\end{figure}
\begin{figure}[h]
\centering
\includegraphics[width=0.5\textwidth]{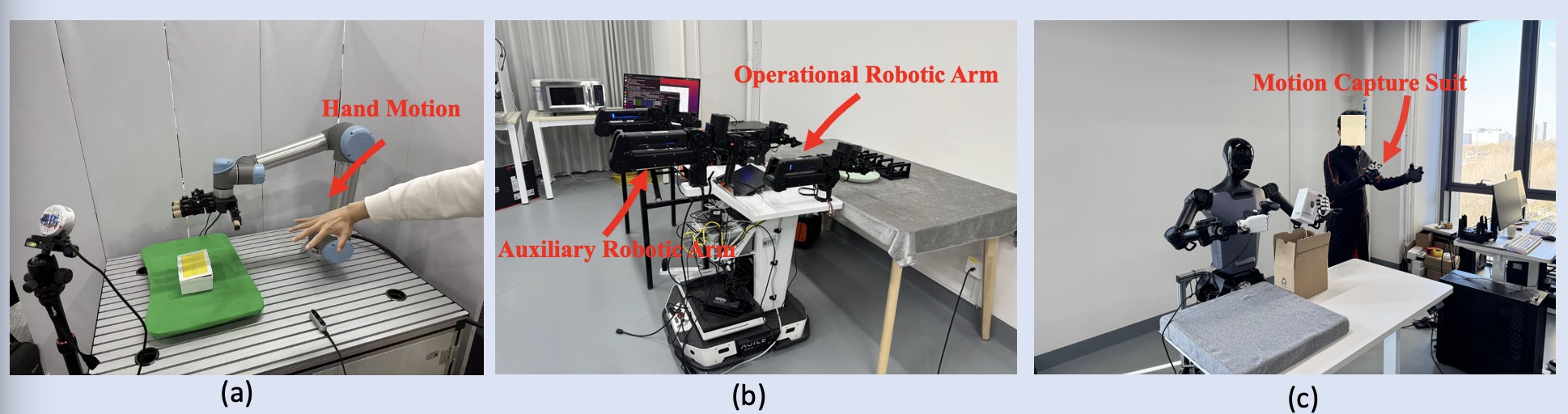}
\caption{Collecting  data methods for different types of robots.}
\label{fig:collect}
\vspace{-2em}
\end{figure}

The visualization of the designed tasks is shown in Figure~\ref{fig:task}. Then, we introduce the data collection process for different robots.
For UR5 single-arm robots with gripper, we use a keyboard interface to control the arm's movements and gripper actions.
For the UR5 single-arm robot with a dexterous hand, we use HaMeR~\citep{pavlakos2024reconstructing} to detect human hand poses with a single RealSense D435 camera. We then employ the AnyTeleop~\citep{qin2023anyteleop} framework to retarget the robot system (see Figure~\ref{fig:collect}(a)).
For the dual-arm robot, Figure~\ref{fig:collect}(b) shows we use an auxiliary robotic arm to control the primary robotic arm to collect the dataset.
For the humanoid robot, we use motion capture suits to map human movements to robot control, enabling the collection of the robot dataset (see Figure~\ref{fig:collect}(c)).
%Figure~\ref{fig:collect} illustrates the visualization for dataset collection methods.
We collect \textbf{50} expert demonstrations utilized for model training.
We conduct \textbf{20} trials for each experiment and report the success rate over these trials to evaluate the performance of \ours.

\noindent\textbf{VLA model experiment setup.}
Evaluating the performance of the VLA model solely based on task success rates is not the only criterion~\cite{zhang2024vlabench}.
Generalization and the ability to understand long-range tasks are critical measures of the effectiveness of the VLA model. Figure~\ref{fig:fvp} shows the four tasks we designed to investigate the spatial understanding, task transfer, language understanding, and long-horizon task performance of the VLA (Vision-Language-Action) model.
These tasks include placing apples at the four corners of the space, picking up bananas and placing them on a plate, pouring water using both arms, and a long-term task that involves placing apples, pouring water, and wiping the table. Each task still requires collecting 50 demos. 

\subsection{Q1:  Can \ours-pretrained policies outperform other imitation learning methods?}

We compare the DP3 and RISE pre-trained by \ours against 2D/3D imitation learning methods on our different robot tasks.
Figure~\ref{tab:combined} shows that \ours pre-training approach can effectively enhance 3D imitation learning such as DP3~\cite{ze2024dp3} and RISE~\cite{wang2024rise}.
Meanwhile, RISE pre-trained by \ours achieves the SOTA performance across these real-world tasks, largely surpassing both 2D and 3D single task imitation learning methods.
Especially in the tasks of dexterous hand, FVP can notably improve the success rate of these tasks, because FVP introduces the time frames to assist visual models in understanding the complexity of motion trajectory on dexterous hand. 

\begin{figure*}[h]
\centering
\includegraphics[width=1\textwidth]{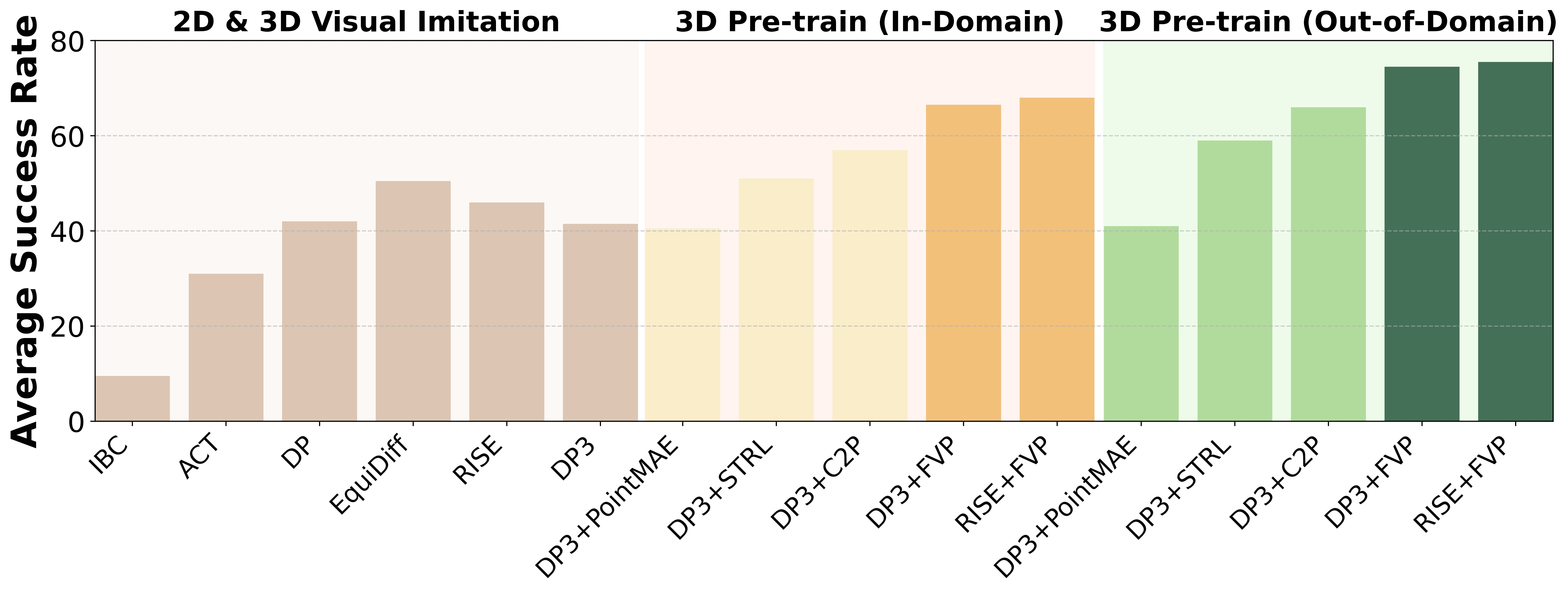}
\vspace{-2em}
\caption{\textbf{Success rate (\%) of imitation learning on real-world robotic tasks and 2D \& 3D visual representations pre-trained by different approaches.} ``DP3+FVP" and ``RISE+FVP" denote the application of \ours to pretrain the visual models from DP3 and RISE, respectively. ``DP3" indicates that the visual model within DP3 has not undergone pretraining. ``DP3+PointMAE", ``DP3+STRL", and ``DP3+C2P" signify the utilization of PointMAE, STRL, and C2P to pre-train the visual model from DP3. The numbers before the comma represent the performance using in-domain datasets for pre-training, while the numbers after the comma represent the performance using out-of-domain datasets for pre-training. }
\label{tab:combined}
\vspace{-1em}
\end{figure*}

\subsection{Q2: Can \ours outperform other pre-trained visual representations?}

We select various 3D/4D pre-training methods (such as PointMAE~\cite{pang2022pointmae},  STRL~\cite{huang2021strl} and C2P~\cite{zhang2023c2p}) to train visual models for comparison with visual models pre-trained by FVP in real-world tasks.
To validate the generalization of the FVP pre-training framework, we pre-train FVP and these baselines using both in-domain and out-of-domain datasets.
For the out-of-domain dataset, we select the Robomind dataset~\cite{wu2024Robomind}, which contains 3D point cloud information.
Figure~\ref{tab:combined}
indicates that whether using an in-domain dataset or an out-of-domain dataset for pre-training, compared to PointMAE~\cite{pang2022pointmae}, STRL~\cite{huang2021strl}, and C2P~\cite{zhang2023c2p}, \ours pre-trained approach can learn more effective visual features, thereby aiding DP3/RISE in improving the more efficacy of real-world robotic task achievement.
Vision encoders pretrained using the Robomind dataset with the FVP framework are considered as general robot vision representations.
%In the supplementary material, we provide the success rates of each task across different methods.
Meanwhile, we compare DP3 pre-trained by \ours with R3M~\citep{nair2022r3m}, MVP~\citep{xiao2022mvp} and MAE (Soup-1M+100 DoH)~\citep{dasari2023unbiased}, which are the large robotic generalized models pre-trained by 2D images. 
We show the performance of using R3M~\citep{nair2022r3m}, MVP~\citep{xiao2022mvp} and MAE (Soup-1M+100 DoH)~\citep{dasari2023unbiased}-trained features in the same policy model as DP3 in Table~\ref{tab:difffusion policy}. We find that FVP pre-training method is more effective in improving the  performance of model on the real-world tasks compared to R3M~\citep{nair2022r3m}, MVP~\citep{xiao2022mvp} and MAE (Soup-1M+100 DoH)~\citep{dasari2023unbiased}.
Similarly to the approach used in R3M~\citep{nair2022r3m}, MVP~\citep{xiao2022mvp} and MAE (Soup-1M+100 DoH)~\citep{dasari2023unbiased}, the DP3 experiment results  in the Table~\ref{tab:difffusion policy}  are also pre-trained using an out-of-domain dataset. Specifically, the visual encoder from DP3 is pre-trained using the Robomind dataset~\cite{wu2024Robomind}.

\begin{table}[h]
\centering
\small
\caption{{\textbf{Success rate (\%) of  2D pre-trained visual representations on the  diffusion policy.} We use the same policy generator as in DP3 to fine-tune R3M, MVP, and MAE (Soup-1M+100 DoH) on the six real-work tasks.} 
}
\vspace{-1em}
\resizebox{0.47\textwidth}{!}{
\begin{tabular}{l!{\vrule width1pt}cccc}
\Xhline{1.5pt}
\multirow{2}{*}{\textbf{}} & \multicolumn{4}{c}{\cellcolor[HTML]{FCF5ED}\textbf{Diffusion Policy for Robotic Action}} \\
&\cellcolor[HTML]{F4BF96}\textbf{R3M}~\citep{nair2022r3m} & \cellcolor[HTML]{F4BF96}\textbf{MVP}~\citep{xiao2022mvp} & \cellcolor[HTML]{F4BF96}\textbf{MAE (Soup-1M+100 DoH)}~\citep{dasari2023unbiased} 
& \cellcolor[HTML]{F4BF96}\textbf{DP3+\ours}\\
\Xhline{1pt}
PickSquarel  & 15/20 & 17/20&18/20 & \cellcolor[HTML]{E0F4FF}\textbf{20/20} \\
PlaceBottle   & 13/20 & 15/20 &15/20 & \cellcolor[HTML]{E0F4FF}\textbf{20/20} \\
\hline
PickPlace  & 14/20 & 16/20&16/20 & \cellcolor[HTML]{E0F4FF}\textbf{17/20} \\
FlipCup  &14/20 &  \cellcolor[HTML]{E0F4FF}\textbf{17/20} &15/20   & 16/20 \\
Assembly   & 9/20 & 10/20 &11/20  & \cellcolor[HTML]{E0F4FF}\textbf{13/20} \\
 ArtiManip   & 11/20 & 14/20 &14/20 & \cellcolor[HTML]{E0F4FF}\textbf{16/20} \\
 \Xhline{1.0pt}
\textbf{Average} & 12.5/20& 15.5/20 &15.3/20  & \cellcolor[HTML]{E0F4FF}\textbf{16.4/20} \\
 \Xhline{1.5pt} 
\end{tabular}}
\vspace{0.05in}
\label{tab:difffusion policy}
\vspace{-1.5em}
\end{table}

\subsection{Q3: Can \ours  improve the effectiveness of VLA models?}
At present, large vision-language-action (VLA) robot models such as RDT-1B~\cite{liu2024rdt} rely on 2D images and robotic proprioceptive data to generate robot actions. Thus, we incorporate a point cloud encoder into the visual component of the original VLA models to support point cloud input. The point cloud visual encoder in the VLA model is the same as the one used in \textbf{iDP3}~\cite{ze2024generalizable}, featuring a pyramid-structured multi-layer fully connected network. 
We group tasks of the same robot type together to fine-tune RDT-1B.
Table~\ref{tab:VLA} shows the performance of RDT-1B, including their versions with point cloud input and pre-trained using FVP, in real-world tasks. We find that incorporating 3D point cloud input and using the FVP pre-training method significantly improves the performance of RDT-1B on real-world tasks.

\begin{table}[h]
\centering
\small
\caption{\textbf{ Success rate (\%) of  five real-world tasks using RDT-1B with different section.} 
``2D Image Input''  and ``3D point cloud Input'' refer to using only images as input and adding point clouds as additional input, respectively.
``2D Image Input by R3M'' and  ``3D encoder pretrained by FVP''  refer to the experimental results using a 2D encoder pretrained with R3M and a 3D encoder pretrained with FVP, respectively, in real-world scenarios.}
\vspace{-1em}
\resizebox{0.48\textwidth}{!}{
\begin{tabular}{l!{\vrule width1pt}ccccc}
\Xhline{1.5pt}
%\multirow{2}{*}{\textbf{Category}} & \multicolumn{8}{c}{\cellcolor[HTML]{FCF5ED}\textbf{Fold Clothes}} \\
\multirow{2}{*}{\textbf{Input Style}} & \multicolumn{5}{c}{\cellcolor[HTML]{FCF5ED}\textbf{RDT-1B~\citep{qi2017pointnet++}}} \\
%\rowcolor{gray!10} % 第二行的颜色
  & \cellcolor[HTML]{F4BF96}\textbf{PickSquarel} &  \cellcolor[HTML]{F4BF96}\textbf{PlaceBottle} &   \cellcolor[HTML]{F4BF96}\textbf{PutBox} &  \cellcolor[HTML]{F4BF96}  \textbf{StackBowl}  &  \cellcolor[HTML]{F4BF96}\textbf{WipePlate}  \\
 \Xhline{1pt}
 2D Image Input  &12/20  &10/20&6/20&8/20&3/20\\
2D Image Input by R3M  &15/20  &12/20&7/20&11/20&4/20\\
 3D point cloud Input &14/20 &12/20 &9/20 &13/20&4/20\\
 3D encoder pretrained by FVP  &\cellcolor[HTML]{E0F4FF}\textbf{18/20} &\cellcolor[HTML]{E0F4FF}\textbf{17/20} &\cellcolor[HTML]{E0F4FF}\textbf{9/20} &\cellcolor[HTML]{E0F4FF}\textbf{16/20} &\cellcolor[HTML]{E0F4FF}\textbf{5/20}\\
 \Xhline{1.5pt}
\end{tabular}}
\label{tab:VLA}
\vspace{-1em}
\end{table}

\begin{table}[h]
\centering
\small
\caption{\textbf{ Success rate (\%) of RDT-1B on the different generalization tasks.} ``FVP'' represents FVP pre-trains the 3D encoder using the Robomind dataset.}
\vspace{-1em}
\resizebox{0.45\textwidth}{!}{
\begin{tabular}{l!{\vrule width1pt}ccc}
\Xhline{1.5pt}
%\multirow{2}{*}{\textbf{Category}} & \multicolumn{8}{c}{\cellcolor[HTML]{FCF5ED}\textbf{Fold Clothes}} \\
\multirow{2}{*}{\textbf{FVP Pre-training}} & \multicolumn{3}{c}{\cellcolor[HTML]{FCF5ED}\textbf{RDT-1B~\citep{qi2017pointnet++}}} \\
%\rowcolor{gray!10} % 第二行的颜色
 & \cellcolor[HTML]{F4BF96}\textbf{2D Image} &  \cellcolor[HTML]{F4BF96}\textbf{3D PointCloud} &   \cellcolor[HTML]{F4BF96}\textbf{FVP}  \\
 \Xhline{1pt}
 Spatial Understanding&8/20  &11/20&\cellcolor[HTML]{E0F4FF}\textbf{14/20}\\
 Knowledge Transfer&10/20  &14/20&\cellcolor[HTML]{E0F4FF}\textbf{16/20}\\
Lanugage Understanding &6/20 &6/20 &\cellcolor[HTML]{E0F4FF}\textbf{7/20} \\
Long Horizon Task &0/20 &2/20 &\cellcolor[HTML]{E0F4FF}\textbf{3/20} \\
 \Xhline{1.0pt}
\textbf{Average}&6/20 &8.25/20 &\cellcolor[HTML]{E0F4FF}\textbf{10/20}\\
\Xhline{1.5pt}
\end{tabular}}
\label{tab:FVP}
\vspace{-1em}
\end{table}

\begin{figure*}[h]
\centering
\includegraphics[width=1.0\textwidth]{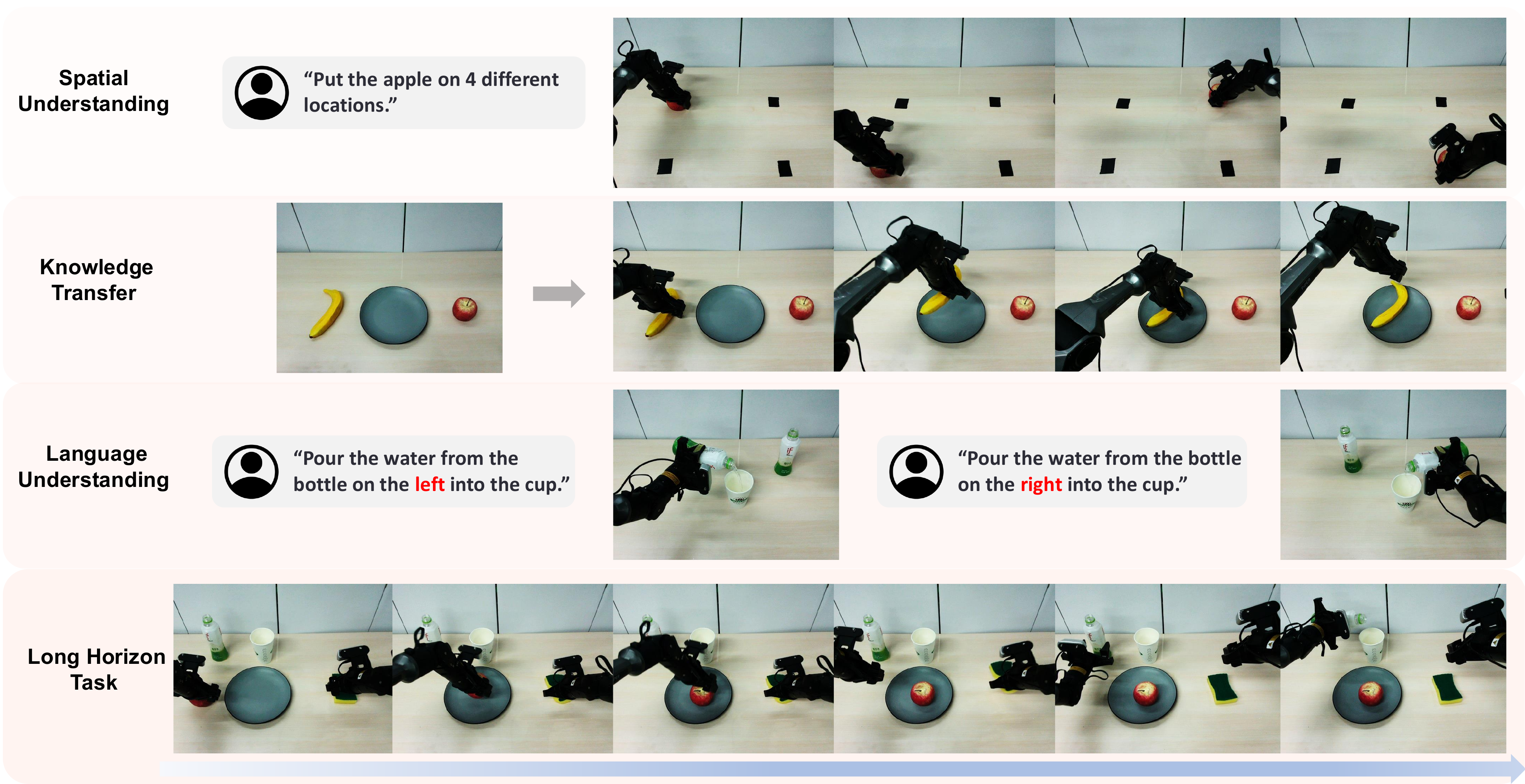}
\vspace{-1em}
\caption{\textbf{ Visualization of the different generalization tasks on RDT-1B.}  We visualize the tasks designed to evaluate various capabilities and generalization of the RDT-1B model.}
\label{fig:fvp}
\vspace{-1em}
\end{figure*}

\subsection{Q4: Can pre-trained VLA exhibit stronger spatial understanding abilities?}

We mainly examine if using 3D point cloud inputs and FVP pre-training can improve the VLA model's spatial perception capabilities. We design a pick-and-place task in which apples are placed in their designated positions based on the given instructions.  We presents the visualization results of the designed tasks in Figure~\ref{fig:fvp}.
Table~\ref{tab:FVP} shows the improvement in spatial perception capabilities of the VLA model with 3D point cloud inputs and FVP pre-training.

\subsection{Q5: Can pre-trained VLA transfer their general knowledge and behavioral abilities to similar but unseen tasks?}
We design a straightforward task in which the model learns to grasp a banana and place it on a plate. Subsequently, we test the model's ability to pick up an apple and place them on the plate, as depicted in the Figure~\ref{fig:fvp}.
From Table~\ref{tab:FVP}, we find that due to the use of a large robotic dataset for pre-training, FVP  can effectively enhance the VLA model's task transferability. Both the 
training and testing language inputs are \textit{“pick up the object from the table and place it on the plate.”}.

\subsection{Q6: Can pre-trained VLA enhanced language understanding ability?}
We aim to verify whether FVP can enhance the robustness of the VLA model in terms of language understanding.
For this purpose, we design an experiment in the same scene where the task is to pour water, with language instructions to control either the left water bottle or the right water bottle to perform the pouring. Figure~\ref{fig:fvp} shows the visualization results of this task. 
During the testing process, we input the language instructions ``Pour the water from the bottle on the \textbf{Left} into the cup " and ``Pour the water from the bottle on the \textbf{Right} into the cup." ten times each. 
Our training set further contains two types of language instructions, 
with an equal number of demonstrations provided for each.
We find that the improvement in language understanding provided by point cloud input to the model is small (see Table~\ref{tab:FVP}).

\subsection{Q7: Can pre-trained VLA accurately support the completion of long-horizon tasks?}

We investigate whether FVP improves performance on long-range tasks. Figure~\ref{fig:fvp} shows the visualization results of a long-horizon task involving multiple dual-arm operations, specifically: placing an apple on a plate, then wiping the table with a sponge, and finally pouring water into a cup.
Table~\ref{tab:FVP} shows that using 3D point cloud input and the FVP pre-training method can effectively enhance the performance of the RDT-1B model on the long-horizon tasks. 

\begin{table}[h]
\caption{\textbf{Ablation study of DP3 pre-trained by \ours on UR5 single arm tasks.} DP3 vision encoder is pre-trained on the Robomind datasets.}
\vspace{-2em}
\small
  \begin{center}
  \resizebox{0.45\textwidth}{!}{
  \begin{tabular}{l!{\vrule width1pt}cccc}
    \Xhline{1.5pt}
    \ & \multicolumn{4}{c}{\cellcolor[HTML]{FCF5ED}\textbf{Real Tasks}}  \\
    \ &  \cellcolor[HTML]{F4BF96}PickSquare  &  \cellcolor[HTML]{F4BF96}PlaceBottle &  \cellcolor[HTML]{F4BF96}PushDraw  &  \cellcolor[HTML]{F4BF96}ToastBread\\
    \Xhline{1pt}
    \textbf{DP3+\ours} &\cellcolor[HTML]{E0F4FF}\textbf{20/20} &\cellcolor[HTML]{E0F4FF}\textbf{20/20} &\cellcolor[HTML]{E0F4FF}\textbf{20/20} &\cellcolor[HTML]{E0F4FF}\textbf{16/20} \\
    \Xhline{1pt}
    %No History Frame &16/20&14/20&15/20&14/20 \\
    Current Frame Input &15/20&14/20 &13/20&13/20\\
    Freeze Visual Encoder &11/20&9/20&10/20&7/20\\
    \Xhline{1.5pt}
  \end{tabular}}
  \end{center}
  \label{tab:ablation}
  \vspace{-1em}
\end{table}

\subsection{Q8: Which components of FVP are important?}
To understand the contributions of each component of \ours, we conduct several ablation studies, as shown in Table~\ref{tab:ablation}.
Specifically, we compare the \textbf{full} \ours with the \textbf{deficient} \ours, which does not history frame point cloud information.
We use the current frame's point cloud instead of the historical frame point cloud to test its impact on FVP performance.
Table~\ref{tab:ablation} shows the 
success rate of  DP3 pre-trained by the full/deficient \ours deployed on the several real-world robotic tasks.
We can find that  the information from historical frames and  have a positive impact on the performance of \ours.
The historical frame information plays a more significant role in the visual representations pre-trained by \ours. 
%We also analyze the impact of using historical frames with different step sizes as the input condition  on FVP's performance. 
%In addition, we also conduct experiments using the current frame's point cloud instead of the historical frame point cloud to test its impact on FVP performance.
%Using the current frame's point cloud as the condition during the pretraining process means that the visual features do not learn the changes in the robot's motion trajectory. 
Table~\ref{tab:ablation}  shows that applying such pre-trained visual features to DP3 does not improve the model's performance.
Finally, we investigate the success rate of downstream tasks when freezing the visual model during the training of DP3. Table~\ref{tab:ablation} shows that freezing the visual model does not lead to an increase in the success rate of real-world tasks. We think this phenomenon is due to the gap between the out-of-domain and in-domain datasets.
We also analyze the impact of using historical frames with different step sizes as the input condition  on FVP's performance. Table~\ref{tab:fvp_performance} demonstrates the performance of FVP when using different historical frame point clouds as inputs in the \textbf{PickSquare} and \textbf{PlaceBottle} task.

\begin{table}[htbp]
    \centering
    \resizebox{0.45\textwidth}{!}{
    \begin{tabular}{lcccc}
        \toprule
        \cellcolor[HTML]{F4BF96}\textbf{Task} & \cellcolor[HTML]{F4BF96}\textbf{1 Frame} & \cellcolor[HTML]{F4BF96}\textbf{2 Frames} & \cellcolor[HTML]{F4BF96}\textbf{3 Frames} & \cellcolor[HTML]{F4BF96}\textbf{4 Frames} \\
        \midrule
        \textbf{PickSquare}  & \cellcolor[HTML]{E0F4FF}\textbf{20/20} & 19/20 & 17/20 & 15/20\\
        \textbf{PlaceBottle} 
        & \cellcolor[HTML]{E0F4FF}\textbf{20/20} & 18/20 & 17/20 & 14/20\\
        \bottomrule
    \end{tabular}
    }
    \caption{Performance of \textbf{DP3+FVP} with Different Historical Frame Point Clouds in the PickSquare and PlaceBottle Tasks}
    \label{tab:fvp_performance}
    \vspace{-1em}
\end{table}

\section{Conclusion}
\vspace{0.5em}
In this work, we introduce 4D Visual Pre-training (\ours), a visual pre-training framework for robotic manipulation, which utilizes the point cloud from history frames and robotic actions to predict the future point clouds as the learning objective, to pre-train a 3D visual representation for downstream robotic tasks. \ours is a general pre-training method for 3D imitation learning methods and we implement \ours upon DP3 and RISE, which results in state-of-the-art results across several real-world manipulation tasks. Additionally, we apply the FVP framework to the VLA (Vision-Language Action) model, which not only improve the success rate of real-world tasks but also enhance the model's generalization capabilities. 

\textbf{Limitations.} Open-source robotics datasets, including Open-X-Embodiment~\cite{2023open}, are available. However, these datasets are missing critical information such as camera parameters and depth data. Thus, we do not utilize these datasets as out-of-domain data for pre-training.

%\clearpage
% The acknowledgments are automatically included only in the final and preprint versions of the paper.

%===============================================================================

{
    \small
    \bibliographystyle{ieeenat_fullname}
    \bibliography{ref}

\begin{thebibliography}{64}
\providecommand{\natexlab}[1]{#1}
\providecommand{\url}[1]{\texttt{#1}}
\expandafter\ifx\csname urlstyle\endcsname\relax
  \providecommand{\doi}[1]{doi: #1}\else
  \providecommand{\doi}{doi: \begingroup \urlstyle{rm}\Url}\fi

\bibitem[Abstreiter et~al.(2021)Abstreiter, Mittal, Bauer, Sch{\"o}lkopf, and Mehrjou]{abstreiter2021diffusionrepresentation}
Korbinian Abstreiter, Sarthak Mittal, Stefan Bauer, Bernhard Sch{\"o}lkopf, and Arash Mehrjou.
\newblock Diffusion-based representation learning.
\newblock \emph{arXiv preprint arXiv:2105.14257}, 2021.

\bibitem[{AgileX Robotics}(2025)]{cobot_magic}
{AgileX Robotics}.
\newblock Cobot magic: An open-source robotic system.
\newblock \url{https://global.agilex.ai/products/cobot-magic}, 2025.
\newblock Accessed: 2025-02-22.

\bibitem[Ajay et~al.(2023)Ajay, Du, Gupta, Tenenbaum, Jaakkola, and Agrawal]{ajay2023conditional}
Anurag Ajay, Yilun Du, Abhi Gupta, Joshua Tenenbaum, Tommi Jaakkola, and Pulkit Agrawal.
\newblock Is conditional generative modeling all you need for decision-making?, 2023.

\bibitem[Chi et~al.(2023)Chi, Feng, Du, Xu, Cousineau, Burchfiel, and Song]{chi2023diffusionpolicy}
Cheng Chi, Siyuan Feng, Yilun Du, Zhenjia Xu, Eric Cousineau, Benjamin Burchfiel, and Shuran Song.
\newblock Diffusion policy: Visuomotor policy learning via action diffusion.
\newblock \emph{arXiv preprint arXiv:2303.04137}, 2023.

\bibitem[Dasari et~al.(2023)Dasari, Srirama, Jain, and Gupta]{dasari2023unbiased}
Sudeep Dasari, Mohan~Kumar Srirama, Unnat Jain, and Abhinav Gupta.
\newblock An unbiased look at datasets for visuo-motor pre-training.
\newblock In \emph{Conference on Robot Learning}, pages 1183--1198. PMLR, 2023.

\bibitem[Deng et~al.(2009)Deng, Dong, Socher, Li, Li, and Fei-Fei]{deng2009imagenet}
Jia Deng, Wei Dong, Richard Socher, Li-Jia Li, Kai Li, and Li Fei-Fei.
\newblock Imagenet: A large-scale hierarchical image database.
\newblock In \emph{2009 IEEE conference on computer vision and pattern recognition}, pages 248--255. Ieee, 2009.

\bibitem[Florence et~al.(2021)Florence, Lynch, Zeng, Ramirez, Wahid, Downs, Wong, Lee, Mordatch, and Tompson]{florence2021implicit}
Pete Florence, Corey Lynch, Andy Zeng, Oscar Ramirez, Ayzaan Wahid, Laura Downs, Adrian Wong, Johnny Lee, Igor Mordatch, and Jonathan Tompson.
\newblock Implicit behavioral cloning, 2021.

\bibitem[Fu et~al.(2024)Fu, Zhao, and Finn]{fu2024mobile}
Z Fu, T~Z Zhao, and C Finn.
\newblock Mobile aloha: Learning bimanual mobile manipulation using low-cost whole-body teleoperation.
\newblock In \emph{8th Annual Conference on Robot Learning (CoRL)}, 2024.

\bibitem[Gervet et~al.(2023)Gervet, Xian, Gkanatsios, and Fragkiadaki]{gervet2023act3d}
Theophile Gervet, Zhou Xian, Nikolaos Gkanatsios, and Katerina Fragkiadaki.
\newblock Act3d: Infinite resolution action detection transformer for robotic manipulation.
\newblock \emph{arXiv preprint arXiv:2306.17817}, 2023.

\bibitem[Goyal et~al.(2023)Goyal, Xu, Guo, Blukis, Chao, and Fox]{goyal2023rvt}
Ankit Goyal, Jie Xu, Yijie Guo, Valts Blukis, Yu-Wei Chao, and Dieter Fox.
\newblock Rvt: Robotic view transformer for 3d object manipulation.
\newblock In \emph{Conference on Robot Learning}, pages 694--710. PMLR, 2023.

\bibitem[Grauman et~al.(2022)Grauman, Westbury, Byrne, Chavis, Furnari, Girdhar, Hamburger, Jiang, Liu, Liu, et~al.]{grauman2022ego4d}
Kristen Grauman, Andrew Westbury, Eugene Byrne, Zachary Chavis, Antonino Furnari, Rohit Girdhar, Jackson Hamburger, Hao Jiang, Miao Liu, Xingyu Liu, et~al.
\newblock Ego4d: Around the world in 3,000 hours of egocentric video.
\newblock In \emph{Proceedings of the IEEE/CVF Conference on Computer Vision and Pattern Recognition}, pages 18995--19012, 2022.

\bibitem[Ho et~al.(2020)Ho, Jain, and Abbeel]{ho2020ddpm}
Jonathan Ho, Ajay Jain, and Pieter Abbeel.
\newblock Denoising diffusion probabilistic models.
\newblock \emph{Advances in neural information processing systems}, 33:\penalty0 6840--6851, 2020.

\bibitem[Huang et~al.(2021)Huang, Xie, Zhu, and Zhu]{huang2021strl}
Siyuan Huang, Yichen Xie, Song-Chun Zhu, and Yixin Zhu.
\newblock Spatio-temporal self-supervised representation learning for 3d point clouds.
\newblock In \emph{Proceedings of the IEEE/CVF International Conference on Computer Vision}, pages 6535--6545, 2021.

\bibitem[Huang et~al.(2023)Huang, Jiang, Ze, and Xu]{huang2023diffusion_reward}
Tao Huang, Guangqi Jiang, Yanjie Ze, and Huazhe Xu.
\newblock Diffusion reward: Learning rewards via conditional video diffusion.
\newblock \emph{arXiv preprint arXiv:2312.14134}, 2023.

\bibitem[Hudson et~al.(2023)Hudson, Zoran, Malinowski, Lampinen, Jaegle, McClelland, Matthey, Hill, and Lerchner]{hudson2023soda}
Drew~A Hudson, Daniel Zoran, Mateusz Malinowski, Andrew~K Lampinen, Andrew Jaegle, James~L McClelland, Loic Matthey, Felix Hill, and Alexander Lerchner.
\newblock Soda: Bottleneck diffusion models for representation learning.
\newblock \emph{arXiv preprint arXiv:2311.17901}, 2023.

\bibitem[Jarzynski(1997)]{jarzynski1997equilibrium}
Christopher Jarzynski.
\newblock Equilibrium free-energy differences from nonequilibrium measurements: A master-equation approach.
\newblock \emph{Physical Review E}, 56\penalty0 (5):\penalty0 5018, 1997.

\bibitem[Liu et~al.(2024)Liu, Wu, Li, Tan, Chen, Wang, Xu, Su, and Zhu]{liu2024rdt}
Songming Liu, Lingxuan Wu, Bangguo Li, Hengkai Tan, Huayu Chen, Zhengyi Wang, Ke Xu, Hang Su, and Jun Zhu.
\newblock Rdt-1b: a diffusion foundation model for bimanual manipulation.
\newblock \emph{arXiv preprint arXiv:2410.07864}, 2024.

\bibitem[Liu et~al.(2019)Liu, Tang, Lin, and Han]{liu2019pvcnn}
Zhijian Liu, Haotian Tang, Yujun Lin, and Song Han.
\newblock Point-voxel cnn for efficient 3d deep learning.
\newblock \emph{Advances in neural information processing systems}, 32, 2019.

\bibitem[Majumdar et~al.(2024)Majumdar, Yadav, Arnaud, Ma, Chen, Silwal, Jain, Berges, Wu, Vakil, et~al.]{majumdar2024vc1}
Arjun Majumdar, Karmesh Yadav, Sergio Arnaud, Jason Ma, Claire Chen, Sneha Silwal, Aryan Jain, Vincent-Pierre Berges, Tingfan Wu, Jay Vakil, et~al.
\newblock Where are we in the search for an artificial visual cortex for embodied intelligence?
\newblock \emph{Advances in Neural Information Processing Systems}, 36, 2024.

\bibitem[Mandlekar et~al.(2021)Mandlekar, Xu, Wong, Nasiriany, Wang, Kulkarni, Fei-Fei, Savarese, Zhu, and Martín-Martín]{mandlekar2021matters}
Ajay Mandlekar, Danfei Xu, Josiah Wong, Soroush Nasiriany, Chen Wang, Rohun Kulkarni, Li Fei-Fei, Silvio Savarese, Yuke Zhu, and Roberto Martín-Martín.
\newblock What matters in learning from offline human demonstrations for robot manipulation, 2021.

\bibitem[Mersch et~al.(2022)Mersch, Chen, Behley, and Stachniss]{mersch2022self}
Benedikt Mersch, Xieyuanli Chen, Jens Behley, and Cyrill Stachniss.
\newblock Self-supervised point cloud prediction using 3d spatio-temporal convolutional networks.
\newblock In \emph{Conference on Robot Learning}, pages 1444--1454. PMLR, 2022.

\bibitem[Nair et~al.(2022)Nair, Rajeswaran, Kumar, Finn, and Gupta]{nair2022r3m}
Suraj Nair, Aravind Rajeswaran, Vikash Kumar, Chelsea Finn, and Abhinav Gupta.
\newblock R3m: A universal visual representation for robot manipulation.
\newblock \emph{arXiv preprint arXiv:2203.12601}, 2022.

\bibitem[Nuti et~al.(2023)Nuti, Franzmeyer, and Henriques]{nuti2023extracting}
Felipe Nuti, Tim Franzmeyer, and João~F. Henriques.
\newblock Extracting reward functions from diffusion models, 2023.

\bibitem[O'Neill et~al.(2023)O'Neill, Rehman, Gupta, Maddukuri, Gupta, Padalkar, Lee, Pooley, Gupta, Mandlekar, et~al.]{2023open}
Abby O'Neill, Abdul Rehman, Abhinav Gupta, Abhiram Maddukuri, Abhishek Gupta, Abhishek Padalkar, Abraham Lee, Acorn Pooley, Agrim Gupta, Ajay Mandlekar, et~al.
\newblock Open x-embodiment: Robotic learning datasets and rt-x models.
\newblock \emph{arXiv preprint arXiv:2310.08864}, 2023.

\bibitem[Pang et~al.(2022)Pang, Wang, Tay, Liu, Tian, and Yuan]{pang2022pointmae}
Yatian Pang, Wenxiao Wang, Francis~EH Tay, Wei Liu, Yonghong Tian, and Li Yuan.
\newblock Masked autoencoders for point cloud self-supervised learning.
\newblock In \emph{European conference on computer vision}, pages 604--621. Springer, 2022.

\bibitem[Pavlakos et~al.(2024)Pavlakos, Shan, Radosavovic, Kanazawa, Fouhey, and Malik]{pavlakos2024reconstructing}
Georgios Pavlakos, Dandan Shan, Ilija Radosavovic, Angjoo Kanazawa, David Fouhey, and Jitendra Malik.
\newblock Reconstructing hands in 3d with transformers.
\newblock In \emph{Proceedings of the IEEE/CVF Conference on Computer Vision and Pattern Recognition}, pages 9826--9836, 2024.

\bibitem[Qi et~al.(2017)Qi, Yi, Su, and Guibas]{qi2017pointnet++}
Charles~Ruizhongtai Qi, Li Yi, Hao Su, and Leonidas~J Guibas.
\newblock Pointnet++: Deep hierarchical feature learning on point sets in a metric space.
\newblock \emph{Advances in neural information processing systems}, 30, 2017.

\bibitem[Qin et~al.(2022)Qin, Wu, Liu, Jiang, Yang, Fu, and Wang]{qin2022dexmv}
Yuzhe Qin, Yueh-Hua Wu, Shaowei Liu, Hanwen Jiang, Ruihan Yang, Yang Fu, and Xiaolong Wang.
\newblock Dexmv: Imitation learning for dexterous manipulation from human videos, 2022.

\bibitem[Qin et~al.(2023)Qin, Yang, Huang, Van~Wyk, Su, Wang, Chao, and Fox]{qin2023anyteleop}
Yuzhe Qin, Wei Yang, Binghao Huang, Karl Van~Wyk, Hao Su, Xiaolong Wang, Yu-Wei Chao, and Dieter Fox.
\newblock Anyteleop: A general vision-based dexterous robot arm-hand teleoperation system.
\newblock \emph{arXiv preprint arXiv:2307.04577}, 2023.

\bibitem[Radosavovic et~al.(2023{\natexlab{a}})Radosavovic, Shi, Fu, Goldberg, Darrell, and Malik]{radosavovic2023robot}
Ilija Radosavovic, Baifeng Shi, Letian Fu, Ken Goldberg, Trevor Darrell, and Jitendra Malik.
\newblock Robot learning with sensorimotor pre-training.
\newblock In \emph{Conference on Robot Learning}, pages 683--693. PMLR, 2023{\natexlab{a}}.

\bibitem[Radosavovic et~al.(2023{\natexlab{b}})Radosavovic, Xiao, James, Abbeel, Malik, and Darrell]{radosavovic2023realmvp}
Ilija Radosavovic, Tete Xiao, Stephen James, Pieter Abbeel, Jitendra Malik, and Trevor Darrell.
\newblock Real-world robot learning with masked visual pre-training.
\newblock In \emph{Conference on Robot Learning}, pages 416--426. PMLR, 2023{\natexlab{b}}.

\bibitem[Rajeswaran et~al.(2017)Rajeswaran, Kumar, Gupta, Vezzani, Schulman, Todorov, and Levine]{rajeswaran2017adroit}
Aravind Rajeswaran, Vikash Kumar, Abhishek Gupta, Giulia Vezzani, John Schulman, Emanuel Todorov, and Sergey Levine.
\newblock Learning complex dexterous manipulation with deep reinforcement learning and demonstrations.
\newblock \emph{arXiv preprint arXiv:1709.10087}, 2017.

\bibitem[Rajeswaran et~al.(2018)Rajeswaran, Kumar, Gupta, Vezzani, Schulman, Todorov, and Levine]{rajeswaran2018learning}
Aravind Rajeswaran, Vikash Kumar, Abhishek Gupta, Giulia Vezzani, John Schulman, Emanuel Todorov, and Sergey Levine.
\newblock Learning complex dexterous manipulation with deep reinforcement learning and demonstrations, 2018.

\bibitem[Rombach et~al.(2022)Rombach, Blattmann, Lorenz, Esser, and Ommer]{rombach2022stable_diffusion}
Robin Rombach, Andreas Blattmann, Dominik Lorenz, Patrick Esser, and Bj{\"o}rn Ommer.
\newblock High-resolution image synthesis with latent diffusion models.
\newblock In \emph{Proceedings of the IEEE/CVF conference on computer vision and pattern recognition}, pages 10684--10695, 2022.

\bibitem[Shafiullah et~al.(2023)Shafiullah, Rai, Etukuru, Liu, Misra, Chintala, and Pinto]{shafiullah2023bringing}
Nur Muhammad~Mahi Shafiullah, Anant Rai, Haritheja Etukuru, Yiqian Liu, Ishan Misra, Soumith Chintala, and Lerrel Pinto.
\newblock On bringing robots home, 2023.

\bibitem[Shah and Kumar(2021)]{shah2021rrl}
Rutav Shah and Vikash Kumar.
\newblock Rrl: Resnet as representation for reinforcement learning.
\newblock \emph{arXiv preprint arXiv:2107.03380}, 2021.

\bibitem[Shridhar et~al.(2023)Shridhar, Manuelli, and Fox]{shridhar2023peract}
Mohit Shridhar, Lucas Manuelli, and Dieter Fox.
\newblock Perceiver-actor: A multi-task transformer for robotic manipulation.
\newblock In \emph{Conference on Robot Learning}, pages 785--799. PMLR, 2023.

\bibitem[Simeonov et~al.(2023)Simeonov, Goyal, Manuelli, Yen-Chen, Sarmiento, Rodriguez, Agrawal, and Fox]{simeonov2023shelving}
Anthony Simeonov, Ankit Goyal, Lucas Manuelli, Lin Yen-Chen, Alina Sarmiento, Alberto Rodriguez, Pulkit Agrawal, and Dieter Fox.
\newblock Shelving, stacking, hanging: Relational pose diffusion for multi-modal rearrangement, 2023.

\bibitem[Song et~al.(2020{\natexlab{a}})Song, Meng, and Ermon]{song2020ddim}
Jiaming Song, Chenlin Meng, and Stefano Ermon.
\newblock Denoising diffusion implicit models.
\newblock \emph{arXiv preprint arXiv:2010.02502}, 2020{\natexlab{a}}.

\bibitem[Song et~al.(2020{\natexlab{b}})Song, Sohl-Dickstein, Kingma, Kumar, Ermon, and Poole]{song2020score}
Yang Song, Jascha Sohl-Dickstein, Diederik~P Kingma, Abhishek Kumar, Stefano Ermon, and Ben Poole.
\newblock Score-based generative modeling through stochastic differential equations.
\newblock \emph{arXiv preprint arXiv:2011.13456}, 2020{\natexlab{b}}.

\bibitem[Sridhar et~al.(2024)Sridhar, Dutta, Jayaraman, Weimer, and Lee]{sridhar2024memoryconsistent}
Kaustubh Sridhar, Souradeep Dutta, Dinesh Jayaraman, James Weimer, and Insup Lee.
\newblock Memory-consistent neural networks for imitation learning, 2024.

\bibitem[Todorov et~al.(2012)Todorov, Erez, and Tassa]{6386109}
Emanuel Todorov, Tom Erez, and Yuval Tassa.
\newblock Mujoco: A physics engine for model-based control.
\newblock In \emph{2012 IEEE/RSJ International Conference on Intelligent Robots and Systems}, pages 5026--5033, 2012.

\bibitem[Urain et~al.(2023)Urain, Funk, Peters, and Chalvatzaki]{urain2023se3diffusionfields}
Julen Urain, Niklas Funk, Jan Peters, and Georgia Chalvatzaki.
\newblock Se(3)-diffusionfields: Learning smooth cost functions for joint grasp and motion optimization through diffusion, 2023.

\bibitem[Wang et~al.(2024{\natexlab{a}})Wang, Fang, Fang, and Lu]{wang2024rise}
Chenxi Wang, Hongjie Fang, Hao-Shu Fang, and Cewu Lu.
\newblock Rise: 3d perception makes real-world robot imitation simple and effective.
\newblock \emph{arXiv preprint arXiv:2404.12281}, 2024{\natexlab{a}}.

\bibitem[Wang et~al.(2024{\natexlab{b}})Wang, Hart, Surovik, Kelestemur, Huang, Zhao, Yeatman, Wang, Walters, and Platt]{wang2024equivariant}
Dian Wang, Stephen Hart, David Surovik, Tarik Kelestemur, Haojie Huang, Haibo Zhao, Mark Yeatman, Jiuguang Wang, Robin Walters, and Robert Platt.
\newblock Equivariant diffusion policy.
\newblock \emph{arXiv preprint arXiv:2407.01812}, 2024{\natexlab{b}}.

\bibitem[Wei et~al.(2023)Wei, Mangalam, Huang, Li, Fan, Xu, Wang, Xie, Yuille, and Feichtenhofer]{wei2023diffusionmae}
Chen Wei, Karttikeya Mangalam, Po-Yao Huang, Yanghao Li, Haoqi Fan, Hu Xu, Huiyu Wang, Cihang Xie, Alan Yuille, and Christoph Feichtenhofer.
\newblock Diffusion models as masked autoencoders.
\newblock In \emph{Proceedings of the IEEE/CVF International Conference on Computer Vision}, pages 16284--16294, 2023.

\bibitem[Wu et~al.(2024)Wu, Hou, Liu, Che, Ju, Yang, Li, Zhao, Xu, Yang, et~al.]{wu2024Robomind}
Kun Wu, Chengkai Hou, Jiaming Liu, Zhengping Che, Xiaozhu Ju, Zhuqin Yang, Meng Li, Yinuo Zhao, Zhiyuan Xu, Guang Yang, et~al.
\newblock Robomind: Benchmark on multi-embodiment intelligence normative data for robot manipulation.
\newblock \emph{arXiv preprint arXiv:2412.13877}, 2024.

\bibitem[{X-Humanoid}(2025)]{x-humanoid}
{X-Humanoid}.
\newblock Tiangong.
\newblock \url{https://x-humanoid.com/bt.html}, 2025.
\newblock Accessed: 2025-03-07.

\bibitem[Xiao et~al.(2022)Xiao, Radosavovic, Darrell, and Malik]{xiao2022mvp}
Tete Xiao, Ilija Radosavovic, Trevor Darrell, and Jitendra Malik.
\newblock Masked visual pre-training for motor control.
\newblock \emph{arXiv preprint arXiv:2203.06173}, 2022.

\bibitem[Yan et~al.(2024)Yan, Wu, and Wang]{yan2024nerfuser}
Ge Yan, Yueh-Hua Wu, and Xiaolong Wang.
\newblock Ne{RF}user: Diffusion guided multi-task 3d policy learning, 2024.

\bibitem[Yang et~al.(2024{\natexlab{a}})Yang, Cao, Deng, Antonova, Song, and Bohg]{yang2024equibot}
Jingyun Yang, Zi-ang Cao, Congyue Deng, Rika Antonova, Shuran Song, and Jeannette Bohg.
\newblock Equibot: Sim (3)-equivariant diffusion policy for generalizable and data efficient learning.
\newblock \emph{arXiv preprint arXiv:2407.01479}, 2024{\natexlab{a}}.

\bibitem[Yang et~al.(2024{\natexlab{b}})Yang, Chen, Sun, and Li]{yang2024visual}
Zetong Yang, Li Chen, Yanan Sun, and Hongyang Li.
\newblock Visual point cloud forecasting enables scalable autonomous driving.
\newblock In \emph{Proceedings of the IEEE/CVF Conference on Computer Vision and Pattern Recognition}, pages 14673--14684, 2024{\natexlab{b}}.

\bibitem[Yu et~al.(2020)Yu, Quillen, He, Julian, Hausman, Finn, and Levine]{yu2020metaworld}
Tianhe Yu, Deirdre Quillen, Zhanpeng He, Ryan Julian, Karol Hausman, Chelsea Finn, and Sergey Levine.
\newblock Meta-world: A benchmark and evaluation for multi-task and meta reinforcement learning.
\newblock In \emph{Conference on robot learning}, pages 1094--1100. PMLR, 2020.

\bibitem[Ze et~al.(2023{\natexlab{a}})Ze, Hansen, Chen, Jain, and Wang]{ze2023vrl3d}
Yanjie Ze, Nicklas Hansen, Yinbo Chen, Mohit Jain, and Xiaolong Wang.
\newblock Visual reinforcement learning with self-supervised 3d representations.
\newblock \emph{IEEE Robotics and Automation Letters}, 8\penalty0 (5):\penalty0 2890--2897, 2023{\natexlab{a}}.

\bibitem[Ze et~al.(2023{\natexlab{b}})Ze, Yan, Wu, Macaluso, Ge, Ye, Hansen, Li, and Wang]{ze2023gnfactor}
Yanjie Ze, Ge Yan, Yueh-Hua Wu, Annabella Macaluso, Yuying Ge, Jianglong Ye, Nicklas Hansen, Li~Erran Li, and Xiaolong Wang.
\newblock Gnfactor: Multi-task real robot learning with generalizable neural feature fields.
\newblock In \emph{Conference on Robot Learning}, pages 284--301. PMLR, 2023{\natexlab{b}}.

\bibitem[Ze et~al.(2024{\natexlab{a}})Ze, Chen, Wang, Chen, He, Yuan, Peng, and Wu]{ze2024generalizable}
Yanjie Ze, Zixuan Chen, Wenhao Wang, Tianyi Chen, Xialin He, Ying Yuan, Xue~Bin Peng, and Jiajun Wu.
\newblock Generalizable humanoid manipulation with improved 3d diffusion policies.
\newblock \emph{arXiv preprint arXiv:2410.10803}, 2024{\natexlab{a}}.

\bibitem[Ze et~al.(2024{\natexlab{b}})Ze, Zhang, Zhang, Hu, Wang, and Xu]{ze2024dp3}
Yanjie Ze, Gu Zhang, Kangning Zhang, Chenyuan Hu, Muhan Wang, and Huazhe Xu.
\newblock 3d diffusion policy: Generalizable visuomotor policy learning via simple 3d representations.
\newblock In \emph{Proceedings of Robotics: Science and Systems (RSS)}, 2024{\natexlab{b}}.

\bibitem[Zhang et~al.(2022)Zhang, Guo, Gao, Fang, Zhao, Wang, Qiao, and Li]{zhang2022pointm2ae}
Renrui Zhang, Ziyu Guo, Peng Gao, Rongyao Fang, Bin Zhao, Dong Wang, Yu Qiao, and Hongsheng Li.
\newblock Point-m2ae: multi-scale masked autoencoders for hierarchical point cloud pre-training.
\newblock \emph{Advances in neural information processing systems}, 35:\penalty0 27061--27074, 2022.

\bibitem[Zhang et~al.(2024)Zhang, Xu, Liu, Yu, Li, Gao, Fei, Yin, Wu, Jiang, et~al.]{zhang2024vlabench}
Shiduo Zhang, Zhe Xu, Peiju Liu, Xiaopeng Yu, Yuan Li, Qinghui Gao, Zhaoye Fei, Zhangyue Yin, Zuxuan Wu, Yu-Gang Jiang, et~al.
\newblock Vlabench: A large-scale benchmark for language-conditioned robotics manipulation with long-horizon reasoning tasks.
\newblock \emph{arXiv preprint arXiv:2412.18194}, 2024.

\bibitem[Zhang et~al.(2023)Zhang, Dong, Liu, and Yi]{zhang2023c2p}
Zhuoyang Zhang, Yuhao Dong, Yunze Liu, and Li Yi.
\newblock Complete-to-partial 4d distillation for self-supervised point cloud sequence representation learning.
\newblock In \emph{Proceedings of the IEEE/CVF conference on computer vision and pattern recognition}, pages 17661--17670, 2023.

\bibitem[Zhao et~al.(2021)Zhao, Jiang, Jia, Torr, and Koltun]{zhao2021point_transformer}
Hengshuang Zhao, Li Jiang, Jiaya Jia, Philip~HS Torr, and Vladlen Koltun.
\newblock Point transformer.
\newblock In \emph{Proceedings of the IEEE/CVF international conference on computer vision}, pages 16259--16268, 2021.

\bibitem[Zhao et~al.(2023)Zhao, Kumar, Levine, and Finn]{zhao2023act}
Tony~Z Zhao, Vikash Kumar, Sergey Levine, and Chelsea Finn.
\newblock Learning fine-grained bimanual manipulation with low-cost hardware.
\newblock \emph{arXiv preprint arXiv:2304.13705}, 2023.

\bibitem[Zheng et~al.(2023)Zheng, Huang, Mei, Hou, Lyu, Dai, Ouyang, and Gong]{zheng2023pointpretrain}
Xiao Zheng, Xiaoshui Huang, Guofeng Mei, Yuenan Hou, Zhaoyang Lyu, Bo Dai, Wanli Ouyang, and Yongshun Gong.
\newblock Point cloud pre-training with diffusion models.
\newblock \emph{arXiv preprint arXiv:2311.14960}, 2023.

\bibitem[Zhou et~al.(2021)Zhou, Du, and Wu]{zhou2021point_voxel_diffusion}
Linqi Zhou, Yilun Du, and Jiajun Wu.
\newblock 3d shape generation and completion through point-voxel diffusion.
\newblock In \emph{Proceedings of the IEEE/CVF international conference on computer vision}, pages 5826--5835, 2021.

\end{thebibliography}
}

\clearpage

\end{document}